\documentclass{article}
\usepackage{cancel}

\usepackage{microtype}
\usepackage{graphicx}
\usepackage{subfigure}
\usepackage{caption}
\usepackage{enumitem}
\usepackage{wrapfig}
\usepackage{scalerel,stackengine}
\stackMath
\newcommand\reallywidehat[1]{%
\savestack{\tmpbox}{\stretchto{%
  \scaleto{%
    \scalerel*[\widthof{\ensuremath{#1}}]{\kern.1pt\mathchar"0362\kern.1pt}%
    {\rule{0ex}{\textheight}}%
  }{\textheight}%
}{2.4ex}}%
\stackon[-6.9pt]{#1}{\tmpbox}%
}
\newcommand{\tx}{\tilde{x}}

\newcommand*\dif{\mathop{}\!\mathrm{d}}
\usepackage{booktabs} %

\usepackage{hyperref}
\usepackage{natbib}

\usepackage[accepted]{icml2024}

\usepackage{amsmath}
\usepackage{amssymb}
\usepackage{mathtools}
\usepackage{amsthm}
\usepackage{cellspace}

\usepackage[capitalize,noabbrev]{cleveref}

\theoremstyle{plain}
\newtheorem{theorem}{Theorem}[section]

\theoremstyle{definition}

\theoremstyle{remark}

\usepackage[textsize=tiny]{todonotes}

\icmltitlerunning{Diffusive Gibbs Sampling}

\begin{document}

\twocolumn[
\icmltitle{Diffusive Gibbs Sampling}

\icmlsetsymbol{equal}{*}

\begin{icmlauthorlist}
\icmlauthor{Wenlin Chen}{equal,cam,mpi}
\icmlauthor{Mingtian Zhang}{equal,ucl}
\icmlauthor{Brooks Paige}{ucl}
\icmlauthor{José Miguel Hernández-Lobato}{cam}
\icmlauthor{David Barber}{ucl}
\end{icmlauthorlist}

\icmlaffiliation{cam}{University of Cambridge, Cambridge, UK}
\icmlaffiliation{mpi}{Max Planck Institute for Intelligent Systems, T\"{u}bingen, Germany}
\icmlaffiliation{ucl}{University College London, London, UK}

\icmlcorrespondingauthor{Wenlin Chen}{wc337@cam.ac.uk}
\icmlcorrespondingauthor{Mingtian Zhang}{m.zhang@cs.ucl.ac.uk}

\icmlkeywords{MCMC, multi-modal sampling, diffusion, Gibbs sampling, molecular dynamics}

\vskip 0.3in
]

\printAffiliationsAndNotice{\icmlEqualContribution} %

\begin{abstract}
The inadequate mixing of conventional Markov Chain Monte Carlo (MCMC) methods for multi-modal distributions presents a significant challenge in practical applications such as Bayesian inference and molecular dynamics. Addressing this, we propose \emph{Diffusive Gibbs Sampling (DiGS)}, an innovative family of sampling methods designed for effective sampling from distributions characterized by distant and disconnected modes. DiGS integrates recent developments in diffusion models, leveraging Gaussian convolution to create an auxiliary noisy distribution that bridges isolated modes in the original space and applying Gibbs sampling to alternately draw samples from both spaces. A novel \emph{Metropolis-within-Gibbs scheme} is proposed to enhance mixing in the denoising sampling step. DiGS exhibits a better mixing property for sampling multi-modal distributions than state-of-the-art methods such as parallel tempering, attaining substantially improved performance across various tasks, including mixtures of Gaussians, Bayesian neural networks and molecular dynamics. 
\end{abstract}

\section{Introduction}

Generating samples from complex unnormalized probability distributions is an important problem in machine learning, statistics and natural sciences. Consider an unnormalized target distribution of the form
\begin{align}
    p(x)=\frac{\exp(-E(x))}{Z},
\end{align}
where $x\in\mathbb{R}^d$ is the variable of interest, $E: \mathbb{R}^d\rightarrow \mathbb{R}$ is a  lower-bounded differentiable energy function, and $Z=\int \exp(-E(x))\dif{x}$ is the (intractable) normalization constant. We aim to draw independent samples $x\sim p(x)$ from the target distribution and estimate expectations of functions $\mathbb{E}_{p(x)}[h(x)]=\int h(x)p(x)\dif{x}$ under the target distribution $p(x)$. 

The gradient of the log density of the target distribution is known as the \emph{score function}:
\begin{equation}
    \nabla_x \log p(x)=-\nabla_x E(x),
\end{equation}
which is independent of $Z$. The score function can be evaluated at any location $x$ since $E$ is assumed to be differentiable. This assumption is commonly satisfied in various practical applications, such as posteriors in Bayesian inference~\citep{welling2011bayesian}, score/energy networks in generative image modelling~\citep{song2019generative}, and Boltzmann distributions in statistical mechanics~\citep{noe2019boltzmann}. Below, we introduce score-based methods for sampling from unnormalized distributions.

\subsection{Score-Based MCMC Methods}\label{sec:score-mcmc}
Unadjusted Langevin Algorithm (ULA)~\citep{grenander1994representations, roberts1996exponential} follows the transition rule given by a discrete-time Langevin SDE:
\begin{equation}
    x_{k+1}=x_k+\eta \nabla_{x} \log p(
x_k)+\sqrt{2\eta}\epsilon_k,
\end{equation}
where $\epsilon_k\sim N(0,I)$. For an infinitesimal step size $\eta$, the Markov chain converges to the target $p(x)$ as $k\to\infty$.

Metropolis-adjusted Langevin Algorithm (MALA)~\citep{roberts1996exponential,roberts2002langevin} defines a proposal $x_{k+1}$ using the ULA update rule and additionally corrects the bias according to the transition probability given by the Metropolis-Hasting (MH) algorithm:
\begin{align}
a_{\text{MALA}}\equiv \min \left\{1, \frac{\exp(-E(x_{k+1}))q(x_k|x_{k+1})}{\exp(-E(x_k))q(x_{k+1}|x_k)}\right\},
\end{align}
where the proposal distribution is given by
\begin{equation}
    q(x'|x)=\mathcal{N}(x'|x+\eta \nabla_x \log p(x), 2\eta I).
\end{equation}
Hamiltonian Monte Carlo (HMC)~\citep{duane1987hybrid,neal2011mcmc}  augments the original variable $x$ with an auxiliary momentum variable $v$, which defines a joint distribution 
\begin{align}
    p(x,v)=p(x)p(v)\propto e^{-E(x)-K(v)},
\end{align}
where $p(v)=\mathcal{N}(v|0,M)$ corresponds to the kinetic energy $K(v)=\frac{1}{2} v^{\text{T}} M^{-1}v$. The total energy, or Hamiltonian, is denoted by $H(x,v)=E(x)+K(v)$. HMC generates samples of $x$ and $v$ by simulating the Hamiltonian equations
\begin{align}
    \frac{\dif x}{\dif t}=\frac{\partial H}{\partial v}=M^{-1}v, ~\text{  }\frac{\dif v}{\dif t}=-\frac{\partial H}{\partial x}=\nabla_x\log p(x).
\end{align}
Accurate numerical simulation of the Hamiltonian equations can be done by the leapfrog algorithm~\citep{neal2011mcmc}, with discretization bias corrected by the MH algorithm.

\label{sec:background}
\begin{figure}[t]
    \centering
\includegraphics[width=0.8\columnwidth]{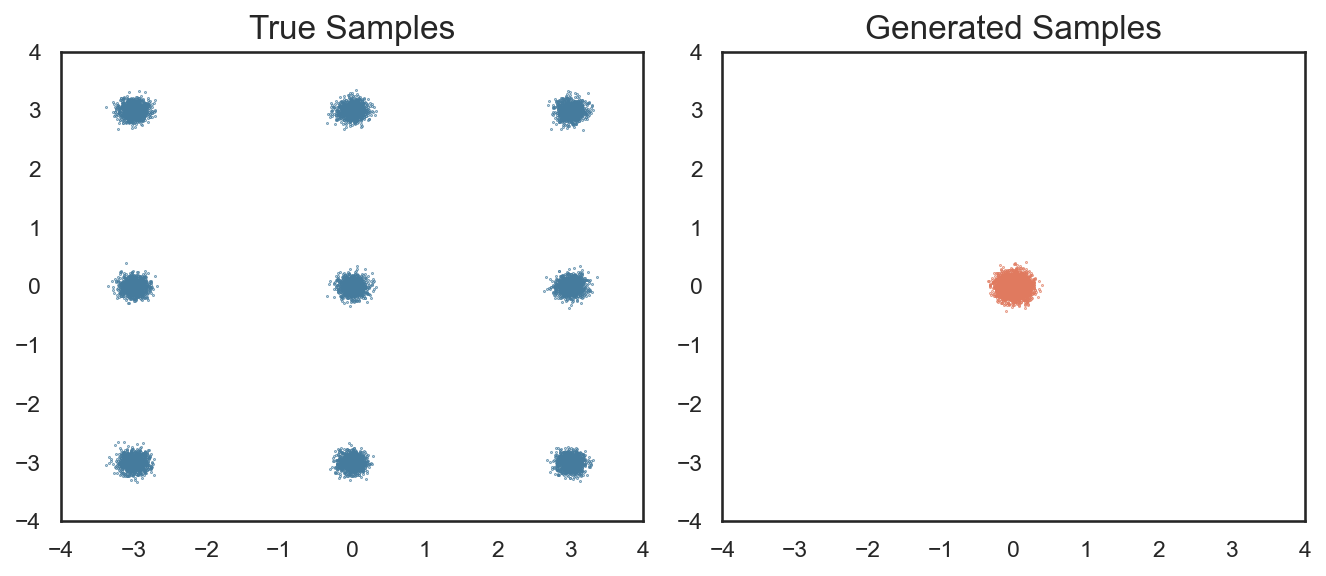}
    \caption{Challenge of multi-modal sampling with score-based MCMC. The true samples represent a mixture of 9 Gaussians and each Gaussian has a standard deviation $\sigma=0.1$. The generated samples are produced by MALA initialized at the origin.}
    \label{fig:mog:mala}
\end{figure}

Despite the effort to address the challenge of exploring the whole support of a distribution using Langevin and Hamiltonian dynamics, MALA and HMC can still be ineffective for distributions with disconnected modes in practice since they often struggle to cross low-density barriers that separate the modes, leading to prolonged transitions from one mode to another~\citep{pompe2020framework}. More generally, score-based sampling methods utilize local gradient information to propose subsequent states, which presents challenges in sampling multi-modal distributions when there is insufficient \emph{bridging density} to connect different modes. This limited connectivity impairs the ability of the resulting samples to accurately represent the entire distribution, a limitation illustrated in Figure \ref{fig:mog:mala}.

\subsection{Convolution-Based Method\label{sec:conv-intro}}
A popular approach to bridging disconnected modes in multi-modal distributions is Gaussian convolution, which has been widely used in the recent developments of diffusion models~\citep{sohl2015deep,song2019generative,ho2020denoising}. For a target distribution $p(x)$ and a Gaussian convolution kernel $p(\tilde{x}|x)=\mathcal{N}(\tilde{x}|\alpha x,\sigma^2I)$, a convolved distribution $p(\tilde{x})$ can be constructed as follows:
\begin{align}
    p(\tilde{x})=\int p(\tilde{x}|x)p(x)\dif{x}.
\end{align}
Since $p(\tilde{x}|x)$ has the full support of $\mathbb{R}^d$, it can effectively create non-negligible density paths between disconnected modes in $p(\tx)$, which makes the modes in  $p(\tx)$ exhibit better connectivity compared to those in $p(x)$.
This cherished property has made Gaussian convolution a popular remedy to heal the blindness of score matching~\citep{song2019generative,wenliang2020blindness,zhang2022towards} and fix KL divergence training for distributions with disjoint or ill-defined density \citep{roth2017stabilizing,zhang2020spread, zhang2023spread,brown2022union}.

Due to the \emph{mode-bridging} property of convolution, it is generally easier for score-based samplers to explore the whole space of $p(\tx)$ than $p(x)$. If we could obtain numerous samples $\tilde{x}\sim p(\tilde{x})$, then it is more likely that these samples will encapsulate a broader range of modes in $p(\tilde{x})$ which are close to different high-density areas in $p(x)$. Consequently, these samples of $\tilde{x}$ can then serve as initial points for sampling from the original target $p(x)$, which facilitates score-based samplers in capturing different modes in $p(x)$. 
However, the score function of the convolved noisy distribution $p(\tx)$ has the form
\begin{align}
\nabla_{\tilde{x}}\log p(\tilde{x})=\nabla_{\tilde{x}} \log \int \exp\left(-E(x)-\frac{\lVert \tilde{x}-\alpha x\rVert^2}{2\sigma^2}\right)\dif{x}\label{eq:intractable:score}
\end{align}
which is typically intractable for non-Gaussian targets. This makes score-based sampling infeasible for $p(\tx)$.

\section{Diffusive Gibbs Sampling}\label{sec:digis-method}
We introduce a novel sampling method, \textbf{Di}ffusive \textbf{G}ibbs \textbf{S}ampling (\emph{DiGS}), which leverages Gaussian convolution with an innovative \emph{Metropolis-within-Gibbs scheme} to enhance multi-modal sampling, while avoiding the intractability of the convolved score function as shown in Equation~\ref{eq:intractable:score}.

\subsection{Sampler Construction}

Instead of trying to directly produce samples from the intractable convolved distribution $p(\tx)$, DiGS employs a Gibbs sampler to sample from the joint distribution $p(x,\tx)=p(\tx|x)p(x)$. This Gibbs sampling procedure involves alternately drawing samples from the two conditional distributions $p(\tx|x)$ (the convolution kernel) and $p(x|\tx)$ (the denoising posterior). In each step, for a given clean sample $x^{(i-1)}$ from the target $p(x)$, we draw
\vspace{-1.8ex}
\begin{enumerate}
    \item a noisy sample $\tx^{(i-1)}\sim p(\tx|x=x^{(i-1)})$,\vspace{-1.2ex}
    \item a new clean sample $x^{(i)}\sim p(x|\tx=\tx^{(i-1)})$.
\end{enumerate}
\vspace{-1.8ex}
For a Gaussian convolution kernel $p(\tx|x)=\mathcal{N}(\tx|\alpha x,\sigma^2I)$,
noisy samples can be easily obtained by corrupting clean samples with Gaussian noises:
\begin{equation}
    \tx = \alpha x + \sigma \epsilon,\quad \epsilon\sim N(0,I),\label{eq:noise corruption}
\end{equation}
where $\alpha\in[0,1]$ is a contraction factor inspired by Diffusion models~\citep{ho2020denoising} and $\sigma$ determines the level of smoothness in the Gaussian convolution. Intuitively, a small $\alpha$ will compress the distribution and make the modes closer, and a large $\sigma$ will encourage the sampler to jump out of the local modes. The effects of these hyperparameters will be investigated in Section~\ref{sec:choose:parameter}, and hyperparameter tuning strategies will be further discussed in Section~\ref{sec:multi-level}.

Unlike the convolved distribution $p(\tx)$ which has an intractable score as shown in Equation~\ref{eq:intractable:score}, the denoising posterior $p(x|\tx)\propto p(x,\tx)=p(\tx|x)p(x)$, taking the form
\begin{align}
  p(x|\tx) \propto \exp\left(-E(x) - \frac{\lVert \alpha x - \tx\rVert^2}{2\sigma^2}\right),\label{eq:posterior}
\end{align}
has a tractable score function~\citep{gao2020learning,RDMC}:
\begin{align}
    \nabla_{x}\log p(x|\tx) = -\nabla_{x} E(x) - \frac{\alpha\left(\alpha x - \tx\right)}{\sigma^2}. \label{eq:posterior:score}
\end{align}
Therefore, common score-based methods like MALA, and HMC could be directly applied to sample from the denoising posterior $p(x|\tx)$ in the denoising sampling step.  It is worth noting that, in contrast to directly sampling from the original target $p(x)\propto \exp(-E(x))$ using score-based methods, incorporating an additional quadratic term as in Equation~\ref{eq:posterior}
improves the Log-Sobolev conditions (i.e., makes it more ``Gaussian-like''), which in turn significantly increases the convergence speed of score-based samplers~\citep{vempala2019rapid};
see~\citet{RDMC} for further in-depth analysis. Below, we show that the DiGS yields a $p(x,\tx)$-irreducible and recurrent Markov Chain under certain regularity conditions.

\begin{figure}[t]
    \centering
    \subfigure[$p(x)$.]{
        \includegraphics[width=0.4\columnwidth]{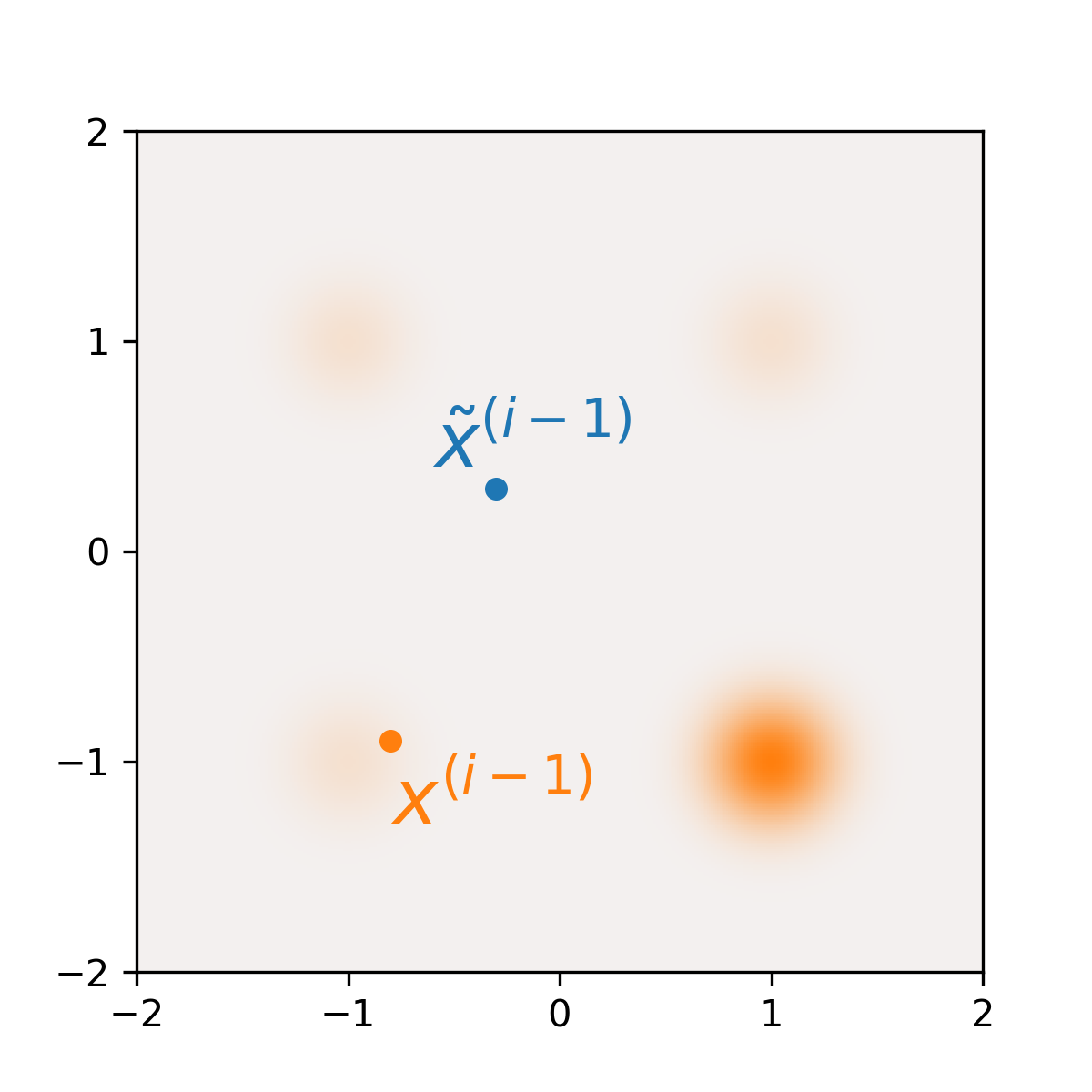}
        \label{fig:unbalanced:mog:density}
    }
    \subfigure[$p(x|\tx^{(i-1)})$.]{\includegraphics[width=0.4\columnwidth]{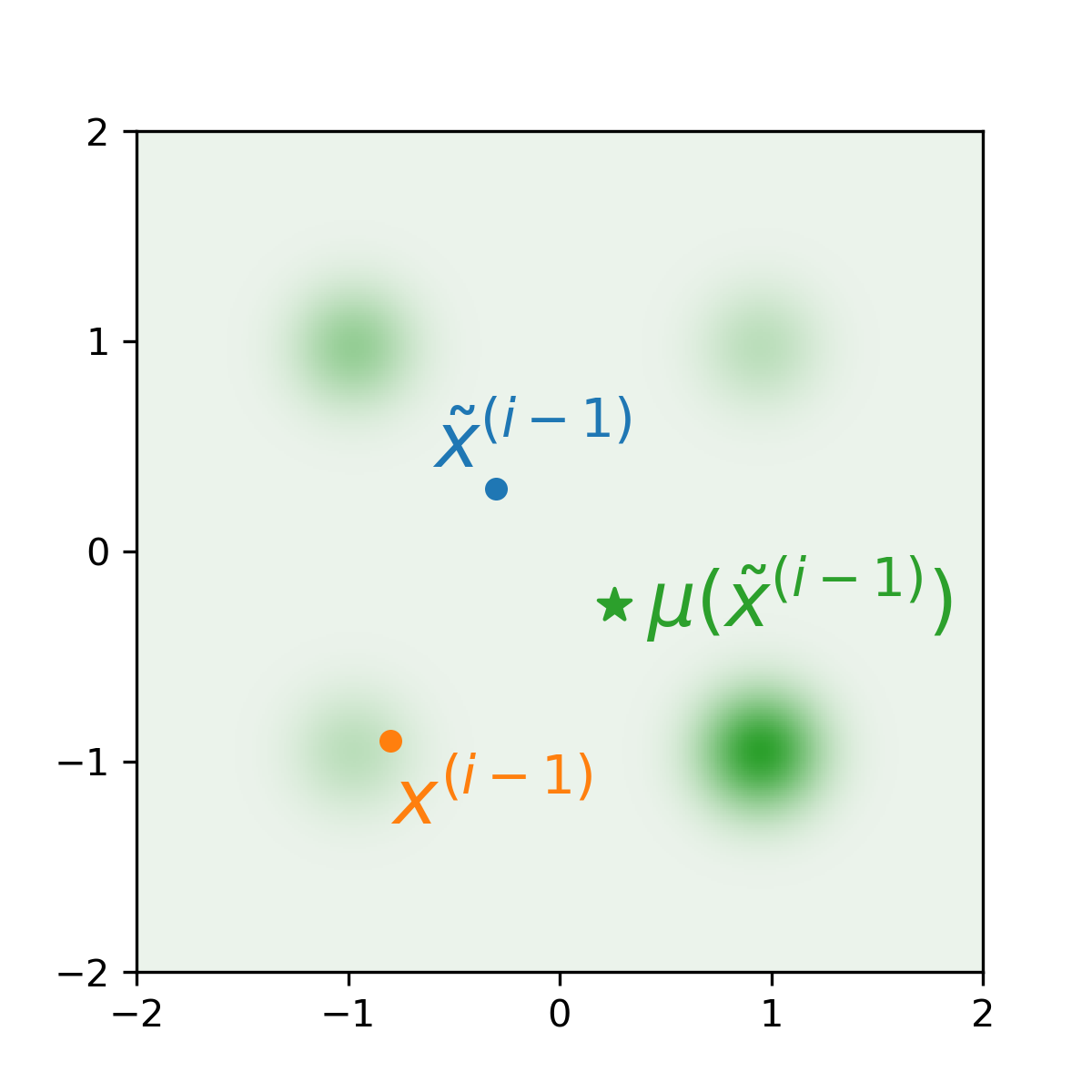}
        \label{fig:unbalanced:mog:posterior}
    }
    \caption{Visualization of an MoG target with unequal weights $w = [0.1,0.1,0.1,0.7]$ for different components. (a) Density heatmap of the target $p(x)$, a clean sample $x^{(i-1)}$ and a noisy sample $\tx^{(i-1)}$. (b) Density heatmap of the denoising posterior $p(x|\tx^{(i-1)})$ with Gaussian convolution parameters $\alpha=1,\sigma=1$. }
    \label{fig:overall-label}
\end{figure}

\begin{theorem}\label{thm:gibbs-valid}
    For an absolutely continuous target distribution $p(x)$, DiGS with a Gaussian convolution kernel $p(\tx|x)=\mathcal{N}(\tx|\alpha x,\sigma^2I)$ ($\alpha>0, \sigma>0$) yields a $p(x,\tx)$-irreducible and recurrent Markov Chain.\label{theorem}
\end{theorem}
The proof of Theorem~\ref{thm:gibbs-valid} can be found in Appendix~\ref{app:proof}. Intuitively, these properties ensure that the chain comprehensively explores the state space from any starting point (irreducibility) and effectively captures the target distribution by infinitely revisiting every state (recurrence)~\citep{robert1999monte}.

In addition to being a valid MCMC sampler, there are some practical considerations that can impact the performance of DiGS. We discuss these in the following sections.

\subsection{Initialization of the Denoising Sampling Step\label{sec:init-denoising-sampling}}

Ideally, one might hope to construct an exact Gibbs sampler where the score-based sampler targeting $p(x|\tx^{(i-1)})$ draws a true sample $x^{(i)}$ at each iteration. Unfortunately, when the target distribution has very disconnected modes, the resulting denoising posterior $p(x|\tx)$ may still exhibit a multi-modal nature. For instance, Figure~\ref{fig:unbalanced:mog:density} illustrates the density of an MoG with unbalanced weights. For a given previous clean sample $x^{(i-1)}$, we generate a noisy sample $\tx^{(i-1)}$ using a Gaussian convolutional kernel with $\alpha=1$ and $\sigma=1$. The corresponding denoising density $p(x|\tx^{(i-1)})$ is depicted in Figure~\ref{fig:unbalanced:mog:posterior}, which exhibits four distinct modes with varying weights. In such scenarios, selecting an appropriate initial point, 
 $ x^{(i)}_{init} $, for the subsequent sampling process $ x^{(i)} \sim p(x|\tx=\tx^{(i-1)}) $ is crucial for score-based samplers.

An ideal initial point for sampling from the denoising posterior $ p(x|\tilde{x}^{(i-1})$ would be the mean of the denoising distribution, defined as $\mu(\tilde{x}) \equiv \int x p(x|\tilde{x})\dif{x}$. By Tweedie's lemma~\citep{efron2011tweedie, robbins1992empirical}, the mean can be expressed as a function of the noisy score $\nabla_{\tilde{x}} \log p(\tilde{x})$:
\begin{align}
\mu(\tilde{x}^{(i-1)}) = \frac{\tilde{x}^{(i-1)} + \sigma^2 \nabla_{\tilde{x}} \log p(\tilde{x}^{(i-1)})}{\alpha}. \label{eq:mean:identity}
\end{align}
Figure~\ref{fig:unbalanced:mog:posterior} demonstrates that the mean function $\mu(\tilde{x}^{(i-1)})$ provides an initial point positioned in the middle of the four modes according to their weights, slightly favoring the mode with the largest weight. However, a challenge arises since $\nabla_{\tilde{x}} \log p(\tilde{x})$ is generally intractable (see Equation~\ref{eq:intractable:score}), making it impractical to compute.

Alternatively, a more straightforward approach is to initialize the denoising sampler at the previous state $x^{(i-1)}$. However, this strategy has a drawback: since $x^{(i-1)}$ is typically close to one of the modes, the score-based sampler often remains trapped in the vicinity of that mode, thereby hindering effective exploration of the entire distribution. Another heuristic initialization strategy is to use the (scaled) noisy sample $\tilde{x}^{(i-1)}/\alpha$, where the scaling factor $\alpha$ reflects the scale relationship between $x$ and $\tilde{x}$ as suggested by the mean function in Equation~\ref{eq:mean:identity}. This approach exhibits a uniform preference for a random mode (for instance, the upper-left mode in Figure~\ref{fig:unbalanced:mog:posterior}) and ignores the underlying weighting between modes, resulting in a bias in representing the true weights of different modes and consequently diminishing the overall quality of the samples.

\begin{figure}[t]
    \centering
        \subfigure[True.]{
            \includegraphics[width=0.22\columnwidth]{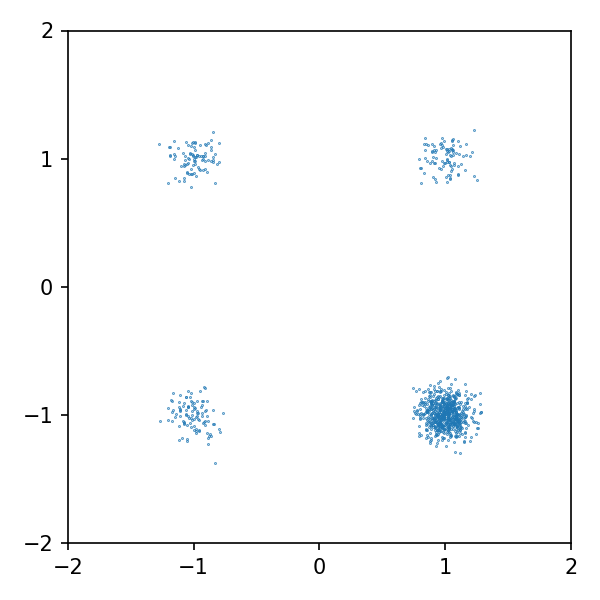}
        }
        \subfigure[$x^{(i-1)}$.]{
    \includegraphics[width=0.22\columnwidth]{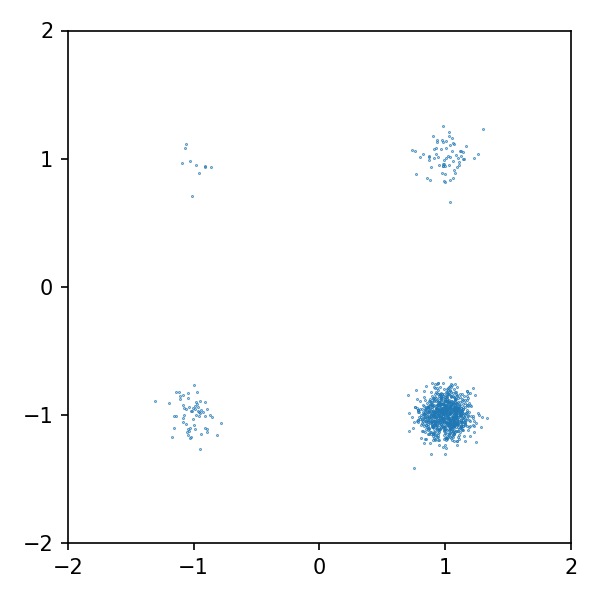}
        }
        \subfigure[$\tx^{(i-1)}/\alpha$.]{
    \includegraphics[width=0.22\columnwidth]{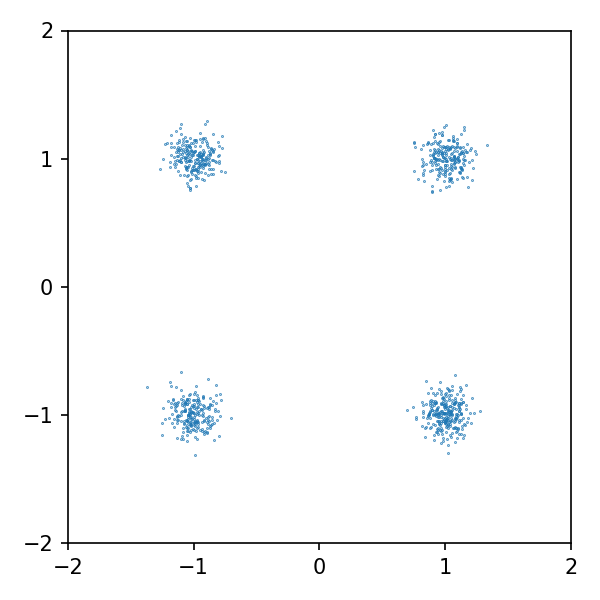}
        }
        \subfigure[\textit{MH}.]{
    \includegraphics[width=0.22\columnwidth]{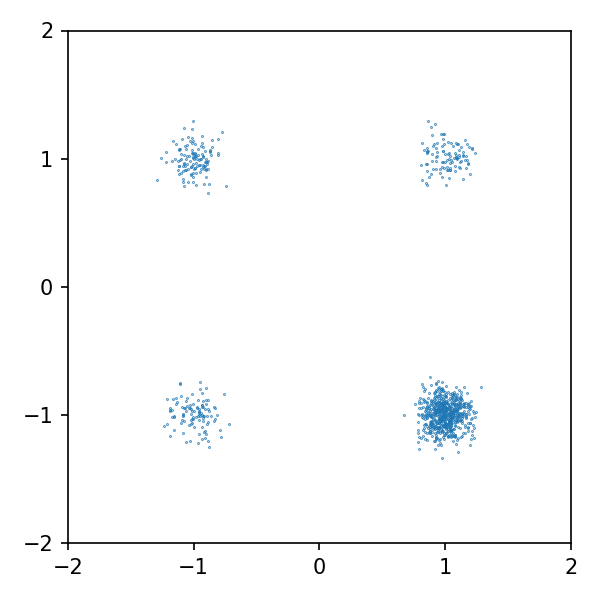}
        }
        
    \caption{Comparison of different initialization techniques for denoising posterior sampling on an unequally weighted MoG target described in Figure~\ref{fig:unbalanced:mog:density}. In each case, we generate 1,000 samples using a Gaussian convolution kernel with $\alpha=1, \sigma=1$.}
    \label{fig:unequal:mog}
\end{figure}
\begin{table}[t]
\vspace{-0.2cm}
        \centering
        \caption{MMD between true samples and samples obtained using different initialization techniques for denoising posterior sampling on an unequally weighted MoG target described in Figure~\ref{fig:unbalanced:mog:density}.} 
        \label{tab:mmd:unequal:mog}
        \begin{tabular}{cccc}
            \hline
            Init. & $x^{(i-1)}$ & $\tx^{(i-1)}/\alpha$ & \textit{MH} \\ \hline
            MMD & $0.15\pm 0.01$ & $0.92\pm 0.04$ & $\mathbf{0.03\pm 0.01}$ \\ 
            \hline
        \end{tabular}
    \end{table}

\begin{figure*}[t]
    \centering
        \subfigure[Effects of $\alpha$ (fix $\sigma=1$).\label{fig:effects:alpha}]{
\includegraphics[width=0.31\linewidth]{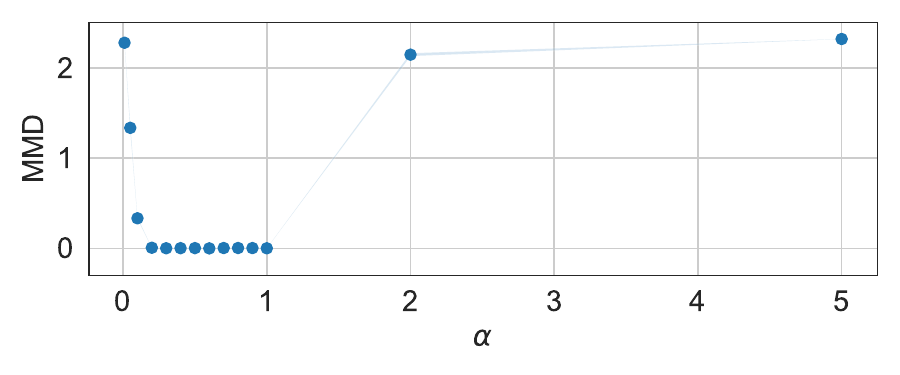}
        }
        \subfigure[Effects of $\sigma$ (fix $\alpha=1$). \label{fig:effects:std}]{
\includegraphics[width=0.31\linewidth]{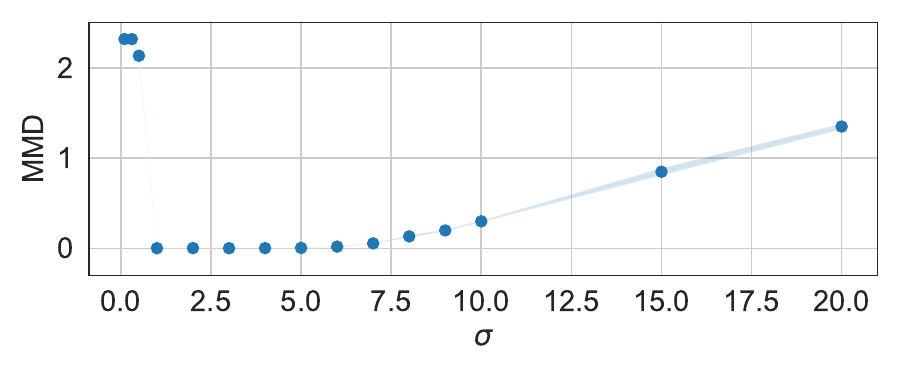}
        }
        \subfigure[Effects of $T$ in the VP schedule.\label{fig:effects:multilevel}]{
\includegraphics[width=0.31\linewidth]{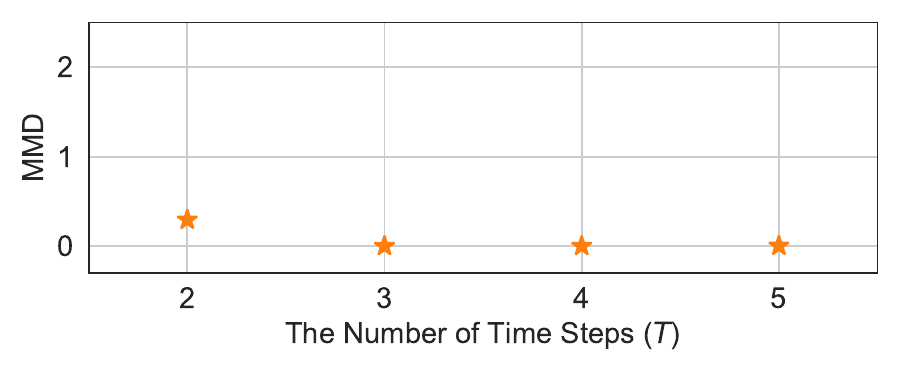}
        }
\caption{Effects of the hyperparameters $\alpha,\sigma$ in Gaussian convolution kernels and the number $T$ of noise levels in the variance-preserving (VP) noise scheduling. The y-axis in all three plots is the MMD between true samples and generated samples generated by DiGS with varying hyperparameters.
Experimental setups can be found in Appendices~\ref{app:conv-param-comparison} and \ref{app:multi-level}.
}
\end{figure*}

\subsubsection{A Metropolis-within-Gibbs Scheme}

To avoid potential bias, we propose a \emph{Metropolis-within-Gibbs scheme} to facilitate mixing across modes by sampling an initialization position $x_{init}'$ from an additional MCMC transition kernel for the subsequent finite score-based sampling steps.
Specifically, to initialize score-based sampling from the denoising posterior $p(x|\tx^{(i-1)})$, we employ a linear Gaussian proposal $x_{init}'\sim q(x|\tx^{(i-1)})$ which is centered at the (scaled) noisy sample $\tx^{(i-1)}/\alpha$:
\begin{equation}
    q(x|\tx^{(i-1)})=\mathcal{N}(x|\tx^{(i-1)}/\alpha,(\sigma/\alpha)^2I),\label{eq:MH proposal}
\end{equation}
where the mean and the variance are chosen according to $q(x|\tx)\propto p(\tx|x)$ inspired by the mean function $\mu(\tx^{(i-1)})$. Note that this proposal only depends on the noisy sample $\tx^{(i-1)}$ and is independent of the previous state $x^{(i-1)}$. We use the MH algorithm to calculate the acceptance rate for this proposal:
\begin{equation}
\small
    a_{init} = \min\left(1, \frac{p(x_{init}'|\tx^{(i-1)})q(x^{(i-1)}|\tx^{(i-1)})}{p(x^{(i-1)}|\tx^{(i-1)})q(x_{init}'|\tx^{(i-1)})}\right),\label{eq:MH:acc:rate}
\end{equation}

where the denoising posterior ratio is tractable since
\begin{align}
\frac{p(x_{init}'|\tx^{(i-1)}))}{p(x^{(i-1)}|\tx^{(i-1)})}
    =\frac{e^{-E(x_{init}')}p(\tx^{(i-1)}|x_{init}')}{e^{-E(x^{(i-1)})}p(\tx^{(i-1)}|x^{(i-1)})}.
\end{align}
If the proposal is accepted, we will initialize the denoising sampling process with the updated value $x_{init}'$, rather than the previous state $x^{(i-1)}$.
Algorithm~\ref{alg:1} summarizes the proposed Diffusion Gibbs Sampling (DiGS) procedure.

To demonstrate the benefits of the additional MCMC kernel updating the initialization, we run DiGS with three different initializations (the previous state $x^{(i-1)}$, a heuristic re-initialization at $\tx^{(i-1)}/\alpha$, and our MH transition strategy)
on an MoG target with different component weights. Figure~\ref{fig:unequal:mog} provides a visual comparison of the samples obtained from these initializations, showing that only the MH transition scheme captures all modes with the correct weightings. For evaluation, we employ the Maximum Mean Discrepancy (MMD)~\citep{gretton2012kernel} throughout all the experiments. The results are shown in Table~\ref{tab:mmd:unequal:mog}, which demonstrates that the MH strategy outperforms the other two in capturing all modes and accurately representing the true weightings. Detailed experimental setup can be found in Appendix~\ref{app:init-comparison}.

\begin{algorithm}[t]
   \caption{Diffusive Gibbs Sampling (DiGS)}
   \label{alg:1}
\begin{algorithmic}[1]
   \STATE {\bfseries Input:} target energy $E(x)$; Gaussian convolution hyperparameters $\alpha,\sigma$; score-based denoising sampler $\mathcal{S}$; the number of denoising sampling steps $L$; the number of Gibbs sampling sweeps $K$; initial clean sample $x^{(0)}$.
   \FOR{$i \gets 1$ to $K$}
   \STATE Draw $\tx^{(i-1)} \sim p(\tilde{x}|x^{(i-1)})$ using Equation~\ref{eq:noise corruption}.
    \STATE Propose $x_{init}'\sim q(x|\tx^{(i-1)})$ as initialization for the denoising process from the proposal in Equation~\ref{eq:MH proposal}.
    \STATE Accept $x_{init}^{(i)}\gets x_{init}'$ with probability $a_{init}$ in Equation~\ref{eq:MH:acc:rate}; otherwise set $x_{init}^{(i)}\gets x^{(i-1)}$.
   \STATE Draw $x^{(i)} \sim p(x|\tilde{x}^{(i-1)})$ by running the score-based sampler $\mathcal{S}$ for $L$ steps from the initial point $x^{(i)}_{init}$ using Equation~\ref{eq:posterior:score}.
   \ENDFOR
   \STATE {\bfseries Output:} $x^{(K)}$
\end{algorithmic}
\end{algorithm}

\subsection{Choosing the Gaussian Convolution Kernels\label{sec:choose:parameter}}
The performance of DiGS can be influenced by the hyperparameters $\alpha$ and $\sigma$ in the Gaussian convolution kernel. Intuitively, for a given fixed $\alpha$, a large $\sigma$ will enhance the exploration capability of the sampler. However, this also makes the denoising posterior $p(x|\tx)$ closer to $p(x)$ as illustrated in Equation~\ref{eq:posterior}, thereby elevating the complexity of denoising sampling.  Similarly, with a fixed $\sigma$, reducing $\alpha$ will bring the modes closer but will also make $\nabla_{x}\log p(x|\tx)$ close to $\nabla_x\log p(x)$ as shown in Equation~\ref{eq:posterior:score}, making the denoising sampling challenging. To illustrate the effects of hyperparameters, we apply DiGS to the MoG problem described in Figure~\ref{fig:mog:mala} with varying values of $\alpha,\sigma$ and show the results in 
Figures~\ref{fig:effects:alpha} and \ref{fig:effects:std}, respectively. This demonstrates that within a specific range of $\alpha$ and $\sigma$ values, DiGS consistently achieves optimal performance indicated by almost zero MMD. However, the quality of samples degrades when hyperparameters deviate beyond certain thresholds.

Although these results suggest that DiGS is robust to a certain range of hyperparameters, selecting an appropriate range remains crucial for each specific target distribution. The optimal range can depend on the support and the shape of the target density, which is typically unknown in practice. Next section will introduce a multi-level noise scheduling to mitigate the need for precise hyperparameter selection.

\subsection{Multi-Level Noise Scheduling~\label{sec:multi-level}}
  Drawing inspiration from the noise scheduling technique used in diffusion models~\citep{ho2020denoising,song2019generative,song2021scorebased,gao2018learning}, we propose to use a sequence of Gaussian convolution kernels with $0<\alpha_T < \cdots < \alpha_1 < 1$ and the corresponding variance $\sigma^2_t=1-\alpha_t^2$ in each Gaussian convolution kernel $p_t(\tx_t|x)=\mathcal{N}(\tx|\alpha_t x,\sigma_t^2I)$. This approach is commonly known as the \emph{variance-preserving} (VP) schedule, where $p_t(\tx_t)=\int p_t(\tx_t|x)p(x)\dif{x}$ and $p_0(\tx_0)\equiv p(x)$. As $\alpha_T\rightarrow 0$, it follows that $p_T(\tx_T|x)\rightarrow\mathcal{N}(\tx_T|0,I)$ and $p_T(\tx_T)\rightarrow \mathcal{N}(\tx_T|0,I)$, which are easy to sample from since they are independent of the characteristics of the target distribution $p(x)$. 
  
  We follow the common sampling procedure in the score-based diffusion model literature~\citep{song2019generative,song2020improved} and propose the following sampling procedure: for each $t$ from $T$ to $1$, we run DiGS to generate a sample, which is used as an initial point in the subsequent time step $t{-}1$.
For a given number of noise levels $T$, we apply a simple linear scheduling scheme to determine the values of $\alpha_t$. Specifically, given the end points $\alpha_1$ and $\alpha_T$, we have
\begin{equation}
    \alpha_t=\alpha_{T}+(\alpha_1-\alpha_{T})\frac{T-t}{T-1},\quad \sigma_t=\sqrt{1-\alpha_t^2}.
\end{equation}
Despite the similarity in noise scheduling to diffusion models, the fundamental sampling mechanism in DiGS is different. In diffusion models, the sampling process requires a progression from time step $T$ to $0$ to yield valid samples. However, DiGS produces a valid sample at any timestep $t$ in principle. This property allows us to set $\alpha_T>0$ and $\alpha_{1}<1$, thereby enhancing the efficiency of DiGS without the necessity to align with the asymptotic distributions. 
To illustrate the effect of the number of noise levels $T$, we implement the VP schedule with a linear noise scheme. We set $\alpha_{T}=0.1, \alpha_{1}=0.9$ and vary $T$ from 2 to 5. 
We test this multi-level DiGS on the MoG problem described in Figure~\ref{fig:mog:mala}. Figure~\ref{fig:effects:multilevel} shows that an optimal sampler is achieved with 
$T>2$ for this problem, circumventing the need for manually selecting Gaussian convolution hyperparameters.

\section{Comparison to Related Methods}\label{sec:connection}
In this section, we explore the relationship between DiGS and related methods, complementing this with empirical comparisons to highlight their distinct characteristics.

\begin{figure}[t]
    \centering
    \includegraphics[width=0.9\columnwidth]{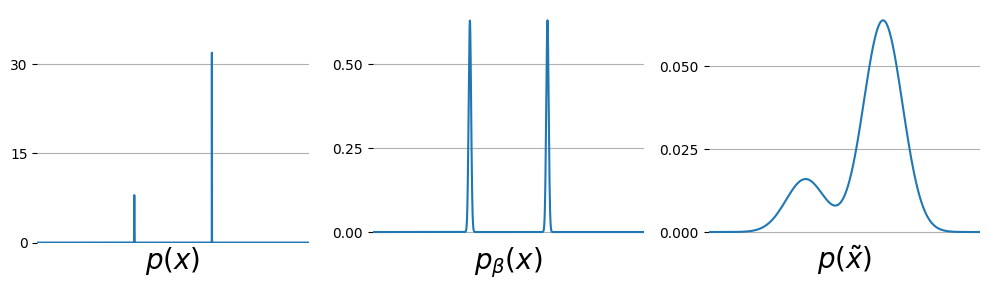}
    \caption{Comparison of a multi-modal target distribution $p(x)$, tempered distribution $p_{\beta}(x)$, and convolved distribution $p(\tx)$.\label{fig:mod-intro}}
\end{figure}

\subsection{Tempering-Based Sampling}\label{sec:tempering-intro}
Tempering-based sampling is a state-of-the-art method for multi-modal target distributions, which samples from smoothed versions of the target distribution and exchanges samples with the original target once in a while. A tempered target distribution is defined as
\begin{align}
    p_\beta(x)\propto p(x)^{\beta}\propto\exp(-\beta E(x)),
\end{align}
where $\beta \equiv 1/\tau<1$ is the inverse temperature. As  $\tau\rightarrow \infty$, the tempered target $p_{\beta\rightarrow 0}(x)$ converges to a flat distribution, which encourages transitions among different modes. Tempering is a key building block to develop state-of-the-art multi-modal sampling methods such as parallel tempering (PT) \citep{swendsen1986replica,geyer1995annealing,syed2022non,surjanovic2022parallel} and annealed importance sampling (AIS) \citep{neal2001annealed}.

Tempering-based sampling makes transitions between distant modes easier. However, tempering is unable to bridge disconnected modes even with a large temperature as it does not alter the support of a distribution, since $p(x)^{\beta}=0$ wherever $p(x)=0$. Moreover, Figure \ref{fig:mod-intro} shows that the tempering-based method is inefficient in connecting modes even when they are not completely disconnected but far away.  Appendix~\ref{sec:comparison-tempering} gives an analytical example, showing the log-density of a point between two modes could approach $-\infty$ in the tempered distribution $p_\beta(x)$, whereas the log-density of that point in the convolved distribution $p(\tx)$ is lower bounded, regardless of the shape of the target $p(x)$.

\begin{figure}[t]
    \centering
        \subfigure[True Samples]{\includegraphics[width=0.3\columnwidth]{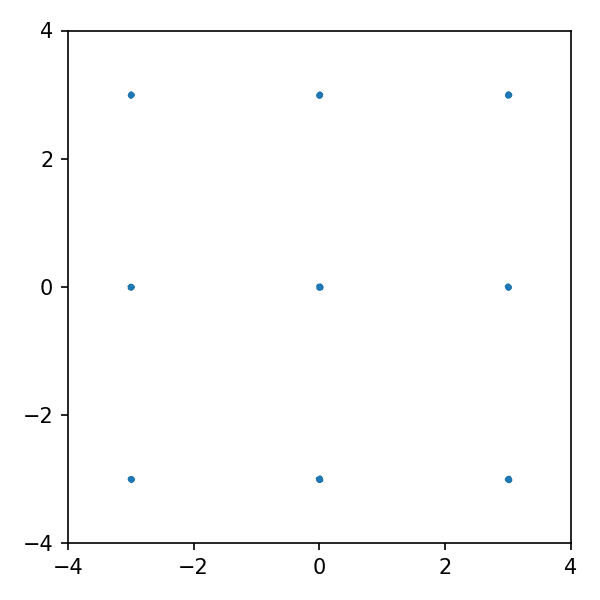}
        }
        \subfigure[PT]{
    \includegraphics[width=0.3\columnwidth]{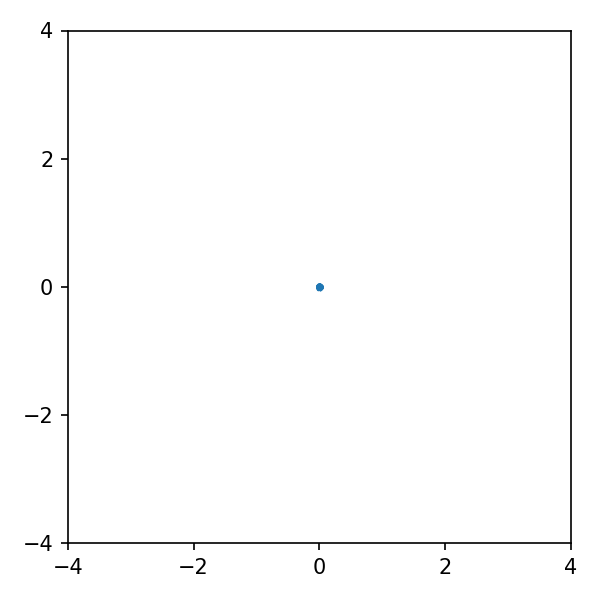}
        }
        \subfigure[\textit{DiGS}]{
    \includegraphics[width=0.3\columnwidth]{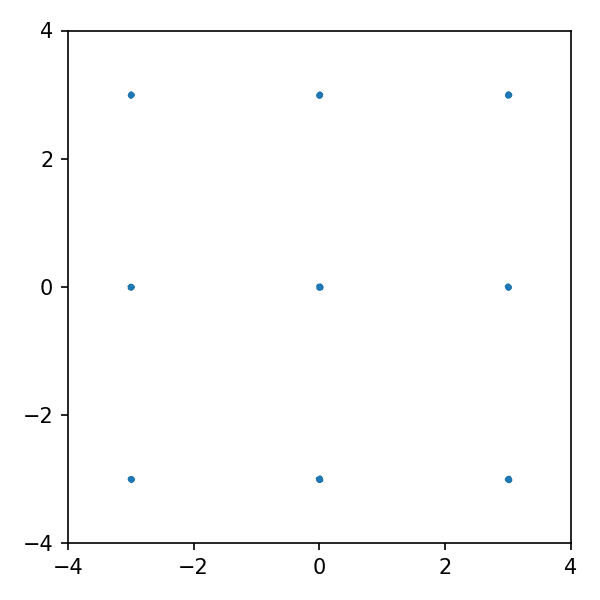}
        }
        \caption{Comparison between parallel tempering (PT) and DiGS on a ``Mixture of Deltas'' problem in 2D with the same budge. PT consists of 5 temperatures, while DiGS uses one noise level. Each sampler is initialized at the origin and generates 1,000 samples. Detailed experimental setup can be found in Appendix~\ref{app:comparison-pt}.\label{fig:mod}}
\end{figure}

To empirically compare the mode-bridging property of PT and DiGS, we conduct experiments on an extreme problem, ``Mixture of Deltas'', which is an MoG with extremely small variance in each Gaussian.
Figure~\ref{fig:mod} shows the comparison of the samples from these two samplers with the same budget: PT is unable to escape the central mode in this extreme setting even with large temperatures $\tau$, whereas DiGS manages to recover all 9 modes, demonstrating its superiority over PT in terms of mode exploration and coverage. To make PT work on this problem, one needs to use more chains with the DEO scheme \citep{deng2023non}; see Appendix~\ref{app:comparison-pt} for a detailed description.

\subsection{Score-Based Diffusion Model}\label{sec:diffusion-models}

The convolution technique plays a key role in score-based diffusion models~\citep{song2019generative,song2021scorebased,song2020improved, ho2020denoising, sohl2015deep,gao2020learning}. These models employ a series of convolutions (indexed by $t$) to form a distribution sequence from $p(\tilde{x}_0)\equiv p(x)$ to a simple distribution $p(\tilde{x}_T)$. At each time step $t$, the score function at time $\nabla_{\tilde{x}_t}\log p(\tilde{x}_t)$ is learned from samples ${x_1,\cdots, x_N}\sim p(x)$ using denoising score matching~\citep{vincent2011connection}. To sample from $p(x)$, ULA and Euler discretization are typically applied in reverse ($t=T\to0$), using samples from each time step $t$ as the initial point at time $t{-}1$. This scheme has shown state-of-the-art generation quality for complex data like images. However, unlike score-based/diffusion generative models where $\nabla_{\tilde{x}}\log p(\tilde{x})$ is learned directly from samples, our setting only assumes access to an energy function $E(x)$ \emph{without} samples, making the noisy marginal score $\nabla_{\tilde{x}}\log p(\tilde{x})$ intractable, as shown in Equation~\ref{eq:intractable:score}.

\citet{zhang2023moment} proposes a pseudo-Gibbs sampler for the joint $p(x,\tx)=p(x)p(\tx|x)$ to obtain clean samples from score-based diffusion models, where $p(\tx|x)=\mathcal{N}(\tx|\alpha x,\sigma^2I)$ and $p(x|\tx)$ is approximated by a full-covariance Gaussian. This approach also requires training data to estimate the score function $\nabla_{\tx}\log p(\tx)$, which is different from our problem setting.

\subsection{Proximal Sampler}
DiGS also belong to the proximal sampler family~\citep{chen2022improved,lee2021structured}, where Gibbs sampling is executed between the Gaussian convolution distribution $p(\tx|x)$ and the denoising posterior $p(x|\tx)$. In the denoising sampling step, proximal samplers find a mode using gradient-based optimization technique and perform rejection sampling around that mode.
We highlight several significant improvements of DiGS over the proximal samplers. First, we employ a score-based sampler such as MALA and HMC in the denoising step instead of rejection sampling around a single mode as in proximal samplers, which helps the method scale to higher-dimensional distributions. Second, we employ an innovative \emph{Metropolis-within-Gibbs scheme} to initialize the denoising sampling step, demonstrating its importance in achieving correct density allocation in multi-modal situations, as discussed in Section~\ref{sec:init-denoising-sampling}.  Third, we emphasize the importance of tuning Gaussian convolution kernel in Section~\ref{sec:multi-level} and employ a multi-level scheduling to eliminate the need of choosing Gaussian convolution hyperparameters. These improvements make the DiGS applicable to practical problems with multi-modal target distributions in high dimensional spaces.

Entropy-MCMC~\citep{li2023entropy} is designed to sample from the flat regions within the posterior of the Bayesian neural network \(p(\theta|\mathcal{D})\), where \(\mathcal{D}\) represents the training dataset. Specifically, the approach begins by defining a surrogate distribution using Gaussian convolution:
\begin{align}
    p(\theta_a|\mathcal{D}) =\int p(\theta_a|\theta)p(\theta|\mathcal{D})\dif \theta,
\end{align}
where \(p(\theta_a|\theta) \propto \exp\left(-\frac{1}{2\eta} \|\theta_a-\theta\|^2_2\right)\) is a Gaussian kernel aimed at \emph{smoothing out} the sharp regions in the posterior. Subsequently, SGLD~\citep{welling2011bayesian} is applied to sample within the joint space of $(\theta, \theta_a)$, where samples of both $\theta$ and $\theta_a$ are kept. It is crucial to highlight that, while this method aims to seek flat modes in the target distribution 
\(p(\theta|\mathcal{D})\), similar to the approaches described in~\citet{chaudhari2019entropy,staines2012variational}, it is not intended and cannot provide exact samples from the target distribution \(p(\theta|\mathcal{D})\) due to its optimization nature. In contrast, DiGS is designed as a valid MCMC sampler that is specifically designed to identify all modes (including both sharp and flat modes) and their corresponding density allocations in the target distribution, thus pursuing a distinct objective.

\begin{figure}[t]
    \centering
\includegraphics[width=0.9\columnwidth]{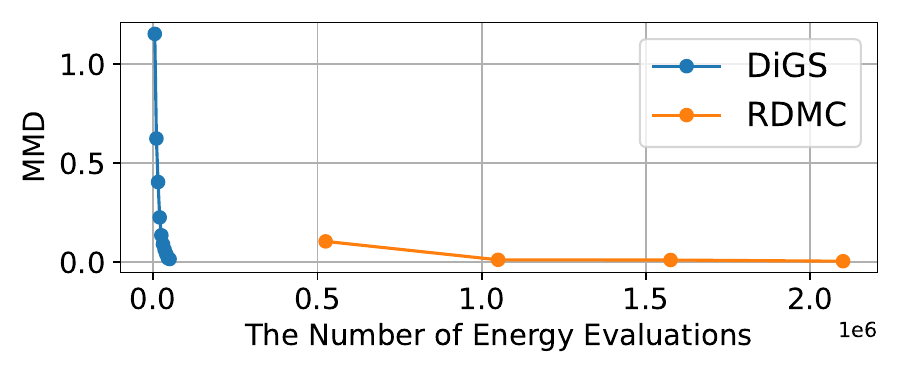}
    \caption{Computational cost comparison between DiGS and RDMC. The x-axis is the number of energy evaluations ($\times10^6$) and the y-axis is the MMD between true and generated samples. We use DiGS with a single noise level $(\alpha=1,\sigma=1)$ and run Gibbs sampling for 1-10 sweeps, represented by the 10 blue scatters. The RDMC is experimented with $T\in\{1,2,3,4\}$, represented by the four orange points in the plot. Detailed experimental setup can be found in Appendix~\ref{app:comparison-rdmc}.\label{fig:digis_rdmc}}
\end{figure}

\begin{figure*}[t]
    \centering
    \subfigure[True Samples]{
        \includegraphics[width=0.183\linewidth]{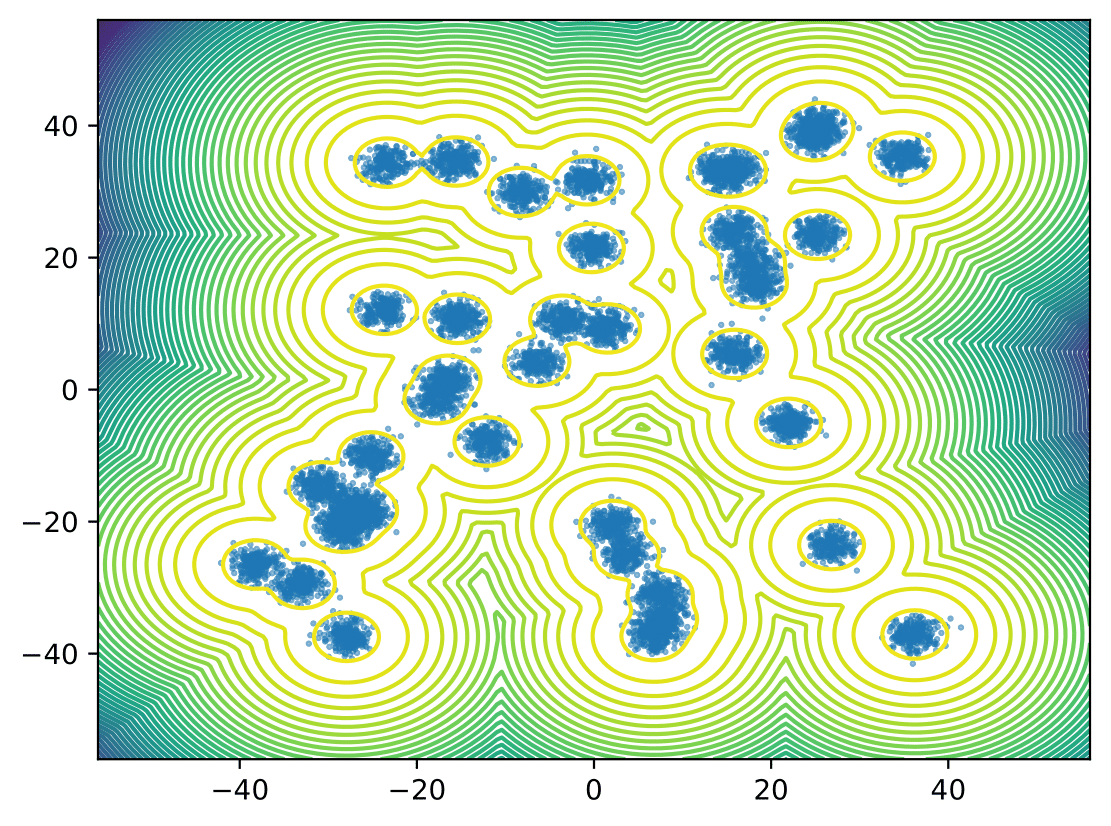}
        \label{fig:mog40-gt}
    }
    \subfigure[MALA]{\includegraphics[width=0.183\linewidth]{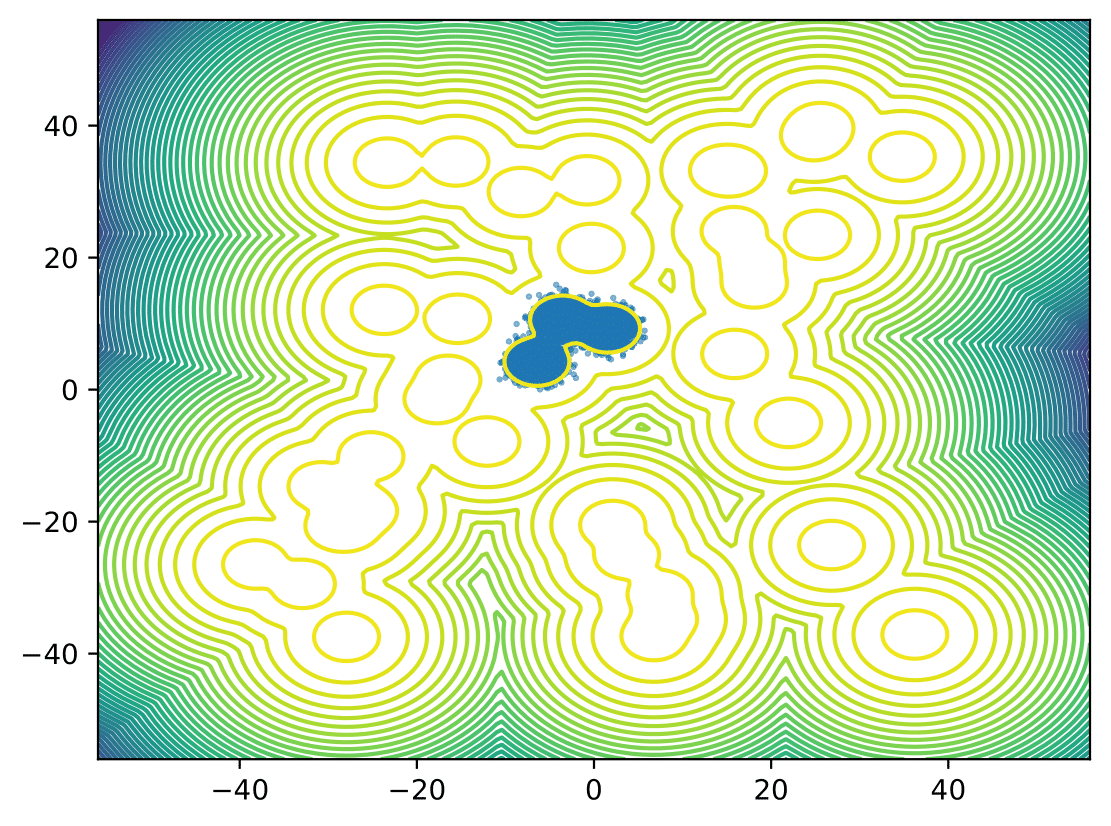}
        \label{fig:mog40-mala}
    }
    \subfigure[HMC]{\includegraphics[width=0.183\linewidth]{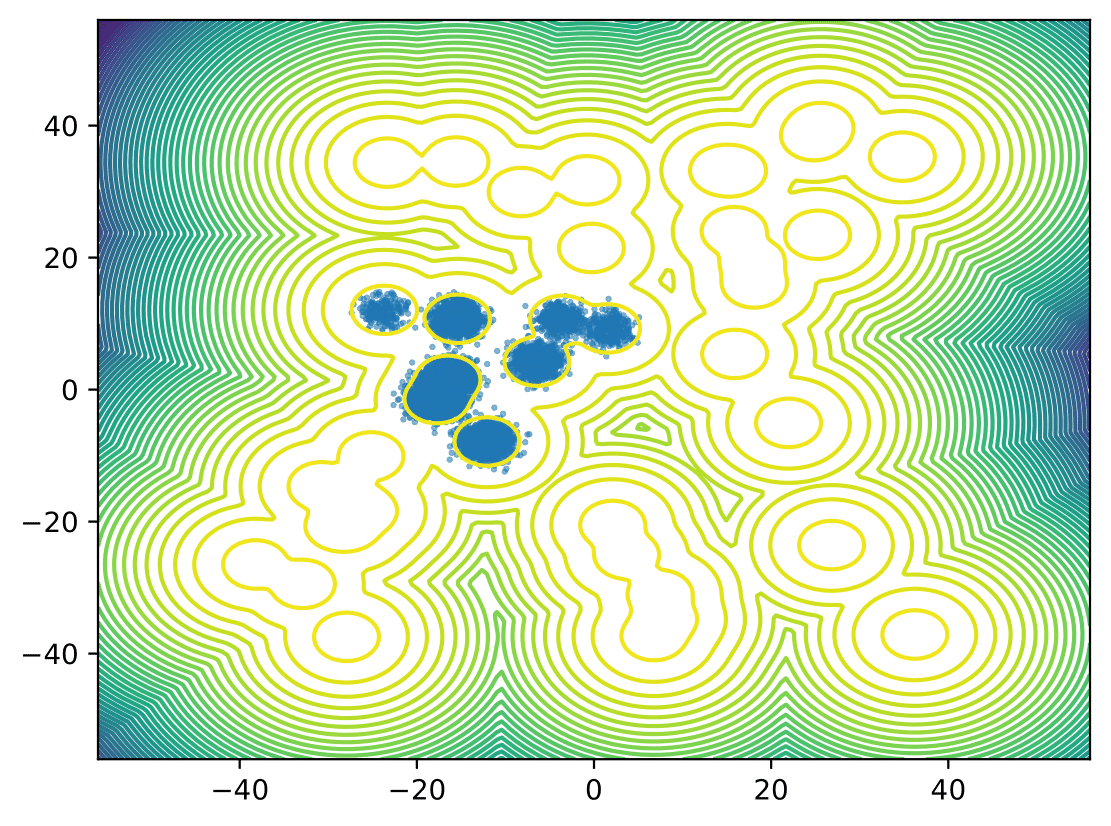}
        \label{fig:mog40-hmc}
    }
    \subfigure[PT]{\includegraphics[width=0.183\linewidth]{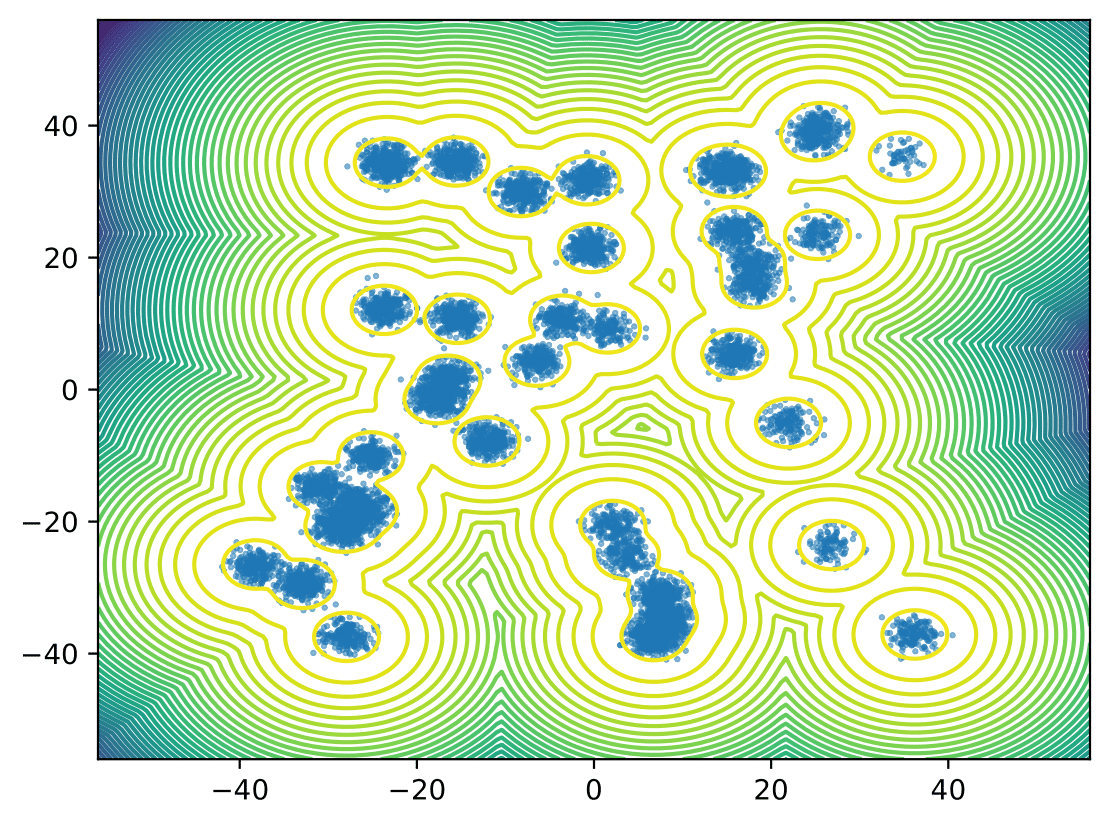}
        \label{fig:mog40-pt}
    }
    \subfigure[\textit{DiGS}]{\includegraphics[width=0.183\linewidth]{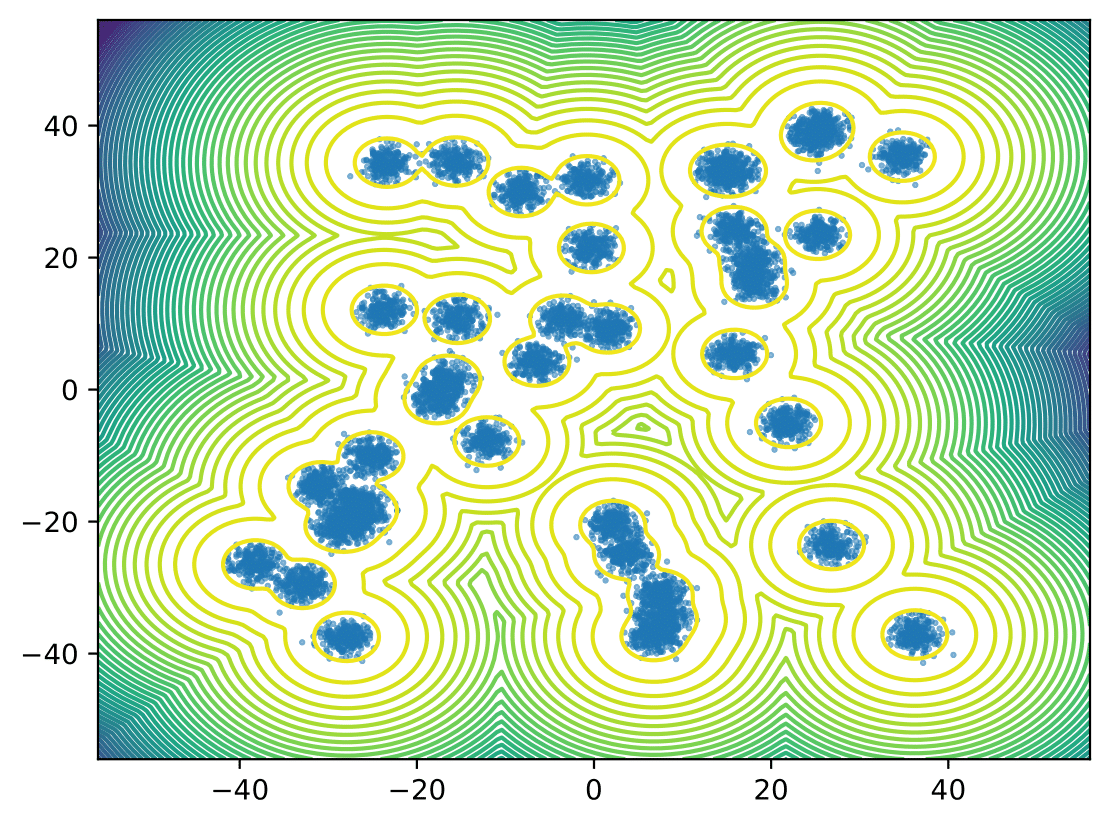}
        \label{fig:mog40-diffgibbs}
    }
    \caption{Visualization of $10^4$ samples for MoG-40 generated by each sampler. All samplers are initialized at the origin.}
    \label{fig:mog40}
\end{figure*}

\subsection{Reverse Diffusion Monte Carlo\label{sec:rdmc}}

To emulate a diffusion model sampling process with access to the unnormalized density of the target only, \citet{RDMC} introduces Reverse Diffusion Monte Carlo (RDMC), which approximates the score function in the noisy space at each time $t$. Specifically, by rewriting the Tweedie's lemma in Equation~\ref{eq:mean:identity}, the noisy score can be expressed as
\begin{align}
    \nabla_{x_t}\log p(x_t)&=\left(\alpha \mu(x_t)-x_t\right)/\sigma_t^2,
\end{align}
where the denoising posterior mean $\mu(x_t)=\int x p(x|x_t)\approx \frac{1}{K}\sum_{k=1}^K x_0^{(t,k)}$ is approximated by the Monte Carlo method,
with samples $x_0^{(t,1)},\cdots, x_0^{(t,K)}$ obtained by running a score-based sampler such as ULA targeting the posterior $p(x_0|x_t)\propto \exp(-E(x_0)-\lVert x_t-\alpha_t x_0\rVert^2/2\sigma_t^2)$. Therefore, in RDMC, obtaining a single sample $x_t \sim p(x_t)$ via ULA at each time step $t\in[1,T]$ involves an intermediate step of generating $K$ posterior samples via another ULA, leading to a \emph{nested} MCMC sampling procedure that imposes substantial computational demands. In contrast, DiGS does not have such hierarchy since it requires only one MCMC chain during the denoising sampling step in each Gibbs sweep, significantly reducing the computational burden.
Figure~\ref{fig:digis_rdmc} compares the computational costs of these two methods when applied to MoG target as shown in Figure~\ref{fig:mog:mala}, demonstrating that DiGS can achieve the same accuracy as RDMC with $10\times$ fewer energy evaluations. The efficiency of DiGS is particularly vital in applications such as molecular configuration sampling (Section~\ref{sec:molecular}), where even a single energy evaluation is costly.

\subsection{Auxiliary Variable MCMC}
DiGS belongs to the broader auxiliary variable MCMC family. This family encompasses various notable methods such as the Swendsen-Wang algorithm~\citep{swendsen1987nonuniversal}, slice sampling~\citep{neal2003slice}, Hamiltonian Monte Carlo (HMC)~\citep{duane1987hybrid,neal2011mcmc}, and auxiliary variational MCMC~\citep{habib2018auxiliary,agakov2004auxiliary}; see~\citet{barber2012bayesian} for a detailed introduction. Among these, HMC bears the closest resemblance to DiGS. We delve into the similarities and differences below.

As introduced in Section~\ref{sec:background}, HMC generates samples from the joint distribution $p(x,v)=p(x)p(v)$ where the $v$ is the auxiliary variable that represents the momentum. The momentum is usually distributed as a Gaussian $p(v) =\mathcal{N}(v|0,\sigma_v^2I)$ that is independent of $x$.  However, there are other variants of HMC where the momentum auxiliary variable $v$ depends on $x$. For example, in Riemannian Manifold HMC~\citep{girolami2011riemann}, the momentum is distributed as $p(v|x) = \mathcal{N}(v|0,\Sigma_v(x))$, where $\Sigma_v(x)$ is the Fisher information matrix that captures the local curvature of the energy around $x$. 
The auxiliary variable in DiGS is the convolved variable $\tx$, which is a noisy version of $x$, given by $p(\tx|x) =\mathcal{N}(\tx|\alpha x,\sigma^2I)$.
Notably, when $\alpha=0$, it follows that $p(\tx|x)=p(\tx) =\mathcal{N}(\tx|0,\sigma^2I)$, and the joint distribution $p(x,\tx)=p(x)p(\tx)$ in DiGS is identical to $p(x,v)$ in the classic HMC formulation. Despite this similarity, DiGS employs Gibbs sampling which alternately samples from the two conditionals $p(\tx|x)$ and $p(x|\tx)$, whereas HMC simulates the Hamiltonian equations as discussed in Section~\ref{sec:score-mcmc}, interleaved with samples from $p(v)$.

\begin{table}[t]
    \footnotesize
    \caption{Sample quality on MoG-40. MMD is computed between true samples and samples generated by each sampler. MAE is computed between the true and estimated expectation values of a quadratic function under the target and is expressed as the percentage of the true expectation value.}
    \label{tab:mog40}
        \centering
        \setlength\extrarowheight{-5pt}
        \setlength{\tabcolsep}{2.5pt}
        \begin{tabular}{cccc}
            \toprule
            Sampler & MMD & MAE (\%) & \#energy eval. \\
            \midrule
            MALA & $1.73\pm0.12$ & $93.3\pm0.73$ & $1.0\times 10^7$ \\
            HMC & $1.70\pm0.09$ & $92.8\pm0.34$ & $1.0\times 10^7$ \\
            PT& $(1.89\pm0.44)\times 10^{-2}$ & $7.32\pm1.85$ & $1.0\times 10^7$ \\
            \textit{DiGS} & $\mathbf{(4.57\pm1.10)\times 10^{-4}}$ & $\mathbf{0.75\pm0.19}$ & $1.0\times 10^7$ \\
            \bottomrule
        \end{tabular}
\end{table}

\section{Empirical Evaluation}\label{sec:experiments}

We evaluate DiGS on three complex multi-modal sampling tasks across various domains\footnote{The code of our experiments can be found in \url{https://github.com/Wenlin-Chen/DiGS}.}: a mixture of 40 Gaussians, Bayesian neural network, and molecular dynamics. For DiGS, we employ MALA with the Metropolis-within-Gibbs scheme. We compare DiGS with three baselines: MALA, HMC and PT. In all experiments, the step sizes of MALA and HMC are tuned via trial-and-error so that the acceptance rates are close to $0.574$~\citep{roberts1998optimal} and $0.65$~\citep{neal2011mcmc}, respectively. We choose not to compare with RDMC, since it is computationally intractable on these complex tasks as demonstrated in Section~\ref{sec:rdmc}.
 
\subsection{Mixture of 40 Gaussians\label{sec:mog-40}}
We first consider a synthetic problem from~\citet{midgley2023flow}, which is a 2D MoG with 40 mixture components. This is a relatively challenging multi-modal sampling task, yet it allows for visual examination of the mode-coverage property for each method.
In this experiment, each method is initialized at the origin and generates $10^4$ samples for evaluation. MALA runs 1,000 Langevin steps per sample. HMC runs 1,000 leapfrog steps per sample. PT consists of 5 chains with temperatures $\tau=\{1.0, 5.62, 31.62, 177.83, 1000.0\}$, where each chain is constructed by an HMC sampler with 200 leapfrog steps per sample. DiGS uses $T=1$ noise level with $\alpha=0.1$ and $\sigma^2=1-\alpha^2$, 200 Gibbs sweeps, and 5 MALA denoising sampling steps per Gibbs sweep.

Figure~\ref{fig:mog40} shows a visual comparison for $10^4$ samples generated by each method. We can see that MALA and HMC fail to explore the modes that are far away from the origin. PT covers all 40 modes but produces significantly less samples for the modes on the top-right and bottom-right corners. DiGS manages to cover all modes with the right amount of samples in each mode. Table~\ref{tab:mog40} shows the Maximum Mean Discrepancy (MMD)~\citep{gretton2012kernel} (computed with 5 kernels with bandwidths $\{2^{-2}, 2^{-1}, 2^0, 2^1, 2^2\}$) between the true samples and samples generated by each sampler and the Mean Absolute Error (MAE) between the true and estimated expectations of a quadratic function under the MoG-40 target. This demonstrates that our method significantly outperforms all baselines on this problem.

\subsection{Bayesian Neural Networks\label{sec:bnn}}

The posterior density of the parameters in a Bayesian neural network (BNN) is known to be complex and multi-modal~\citep{barber1998ensemble,hernandez2015probabilistic, louizos2017multiplicative,izmailov2018averaging}. For a given training dataset $D_{\text{train}}=\{(x_i,y_i)\}_{i=1}^N$, the posterior density can be expressed as $p(\theta|D_{\text{train}})\propto p(\theta)\prod_i p(y_i|x_i,\theta)$, where $p(\theta)$ is the prior density over the parameters and $p(y|x,\theta)$ is the likelihood given by the NN $f_{\theta}(x)$ for a data point $(x,y)$. We consider a three-layer neural network with ReLU activation, input-layer size $d_x=20$, hidden-layer size $d_h=25$, and output-layer size $d_y=1$. This results in $d=550$ parameters in total. We use a Gaussian prior $p(\theta)=\mathcal{N}(\theta|0,\sigma_p^2 I)$ with $\sigma_p=1/\sqrt{d_{\text{fan-in}}}$ and a Gaussian likelihood $p(y|x,\theta)=\mathcal{N}(y|f_{\theta}(x),\sigma_n^2)$ with $\sigma_n=0.1$. We sample the ground-truth parameters $\theta^*\sim p(\theta)$ from the prior and use $\theta^*$ to generate $N=500$ training points and 500 test points for evaluation. 

All methods are initialized at the same random sample from the prior and generate $150$ samples from the posterior $p(\theta|D_{\text{train}})$ for evaluation. MAP (Maximum a Posteri) runs $7.5\times 10^5$ full-batch gradient descent steps. MALA runs 5,000 Langevin steps per sample. HMC runs 5,000 leapfrog steps per sample. PT consists of 5 chains with temperatures $\tau=\{1.0, 5.62, 31.62, 177.83, 1000.0\}$, where each chain is constructed by an HMC sampler with 1,000 leapfrog steps per sample. DiGS uses the VP schedule with $T=5$ noise levels, ranging from $\alpha_T=0.1$ to $\alpha_1=0.9$, each with 100 Gibbs sweeps and 10 MALA denoising sampling steps per Gibbs sweep. 
Table~\ref{tab:bnn} reports the average predictive negative log-likelihood (NLL) for each method on the test data, which shows that DiGS significantly outperforms other baselines. We speculate that the performance gain comes from the fact that DiGS captures a broader range of modes.

\begin{table}[t]
    \normalsize
    \caption{Average test predictive NLL for the BNN estimated by $10^3$ samples generated by each sampler.}
    \label{tab:bnn}
        \centering
        \setlength\extrarowheight{-5pt}
        \begin{tabular}{ccc}
            \toprule
            Sampler & NLL & \#energy evaluations \\
            \midrule
            MAP & $0.548\pm0.066$ & $5.0\times10^6$ \\
            MALA & $0.399\pm0.014$ & $5.0\times10^6$ \\
            HMC & $0.315\pm0.012$ & $5.0\times10^6$ \\
            PT & $0.241\pm0.005$ & $5.0\times10^6$ \\
            \textit{DiGS} & $\mathbf{0.189\pm0.002}$ & $5.0\times10^6$  \\
            \bottomrule
        \end{tabular}
\end{table}
\begin{table*}[t]
\centering
\small
\begin{minipage}{0.9\textwidth}
    \caption{KL divergences for the Ramachandran plots and the marginals of the dihedral angles $\phi$ and $\psi$ in alanine dipeptide.}
    \label{tab:molecule}
        \centering
        \setlength\extrarowheight{-5pt}
        \begin{tabular}{ccccccc}
            \toprule
            Sampler & $p(\phi)$ & $p(\psi)$ & Ramachandran & \#energy evaluations & \#samples \\
            \midrule
            MALA  & $1.9{\times}10^{-3}$ & $5.3{\times}10^{-4}$ & $1.1{\times} 10^{-2}$ & $1.0{\times} 10^9$ & $10^6$ \\
            HMC & $4.0{\times}10^{-4}$ & $3.1{\times}10^{-4}$ & $7.1{\times}10^{-3}$ & $1.0{\times} 10^9$ & $10^6$ \\
            \textit{DiGS} & $\mathbf{2.3{\times}10^{-4}}$ & $\mathbf{1.2{\times} 10^{-4}}$ & $\mathbf{4.3{\times} 10^{-3}}$ & $1.0{\times} 10^9$ & $10^6$ \\
            \midrule
            MD PT \citep{midgley2023flow} & $2.5{\times}10^{-4}$ & $1.4{\times} 10^{-4}$ & $5.3{\times} 10^{-3}$ & $2.3{\times}10^{10}$ & $10^6$\\
            Ground-truth \citep{midgley2023flow} & -- & -- & -- & $2.3{\times}10^{11}$ & $10^7$ \\
            \bottomrule
        \end{tabular}
        \end{minipage}
\end{table*}

\subsection{Molecular Configuration Sampling\label{sec:molecular}}

Finally, we consider a real-world problem of sampling equilibrium molecular configurations from the Boltzmann distribution of the 22-atom molecule alanine dipeptide in an implicit solvent at temperature 300K, where the potential energy $E(x)$ is a function of 3D atomic coordinates obtained by simulating physical laws~\citep{wu2020stochastic,dibak2022temperature,campbell2021gradient,stimper2022resampling,midgley2023flow}. This is a very challenging problem since $E(x)$ is highly multi-modal with many high energy barriers and is also costly to evaluate. Following the setup in~\citet{midgley2023flow}, we represent the molecule with $d=60$ roto-translation invariant internal coordinates.

\begin{wrapfigure}{r}{0.43\columnwidth} %
    \centering
    \vspace{-0.3cm}\includegraphics[width=0.43\columnwidth]{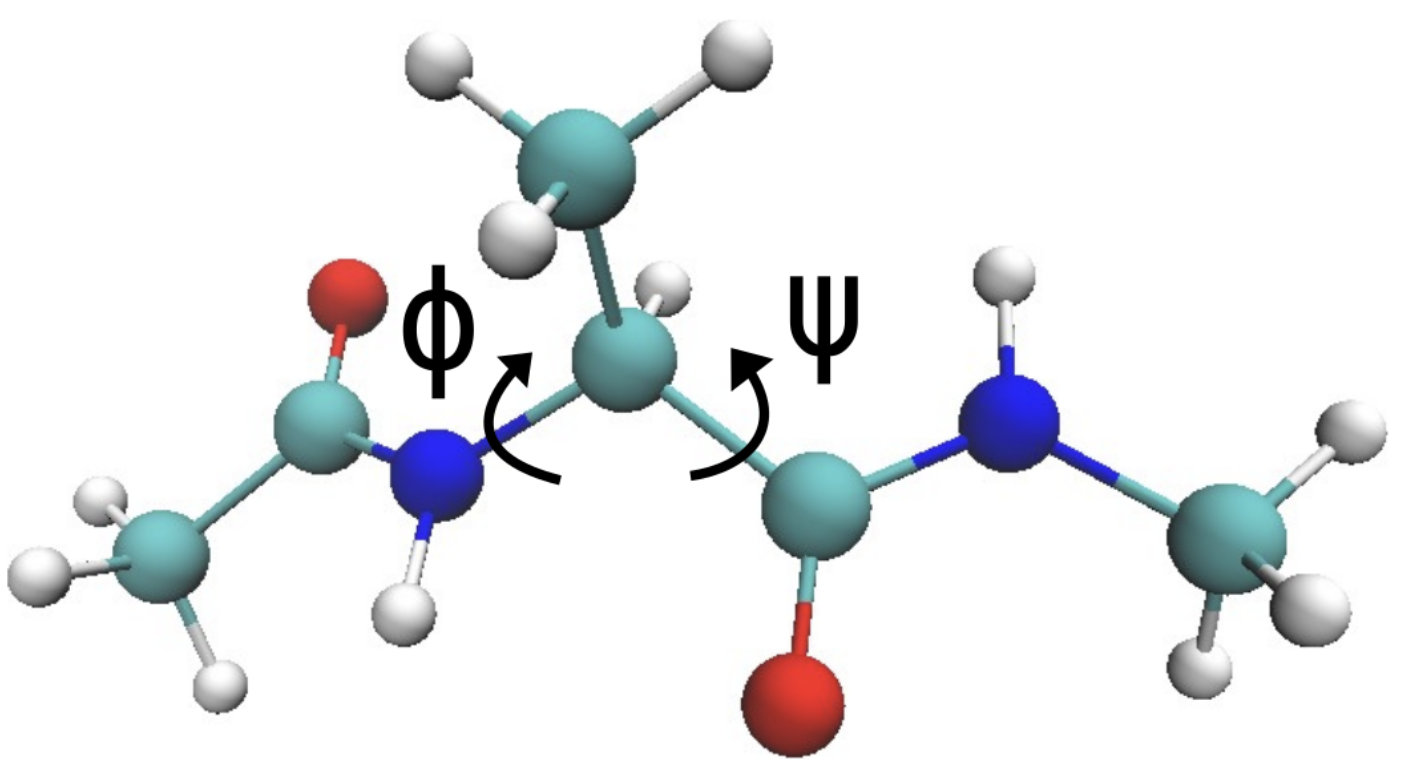}
    \caption{Alanine dipeptide \citep{midgley2023flow}}
    \label{fig:aldp-visualization}
\end{wrapfigure}
\begin{figure}[t]
    \centering
    \subfigure[Ground-truth]{\includegraphics[width=0.45\columnwidth]{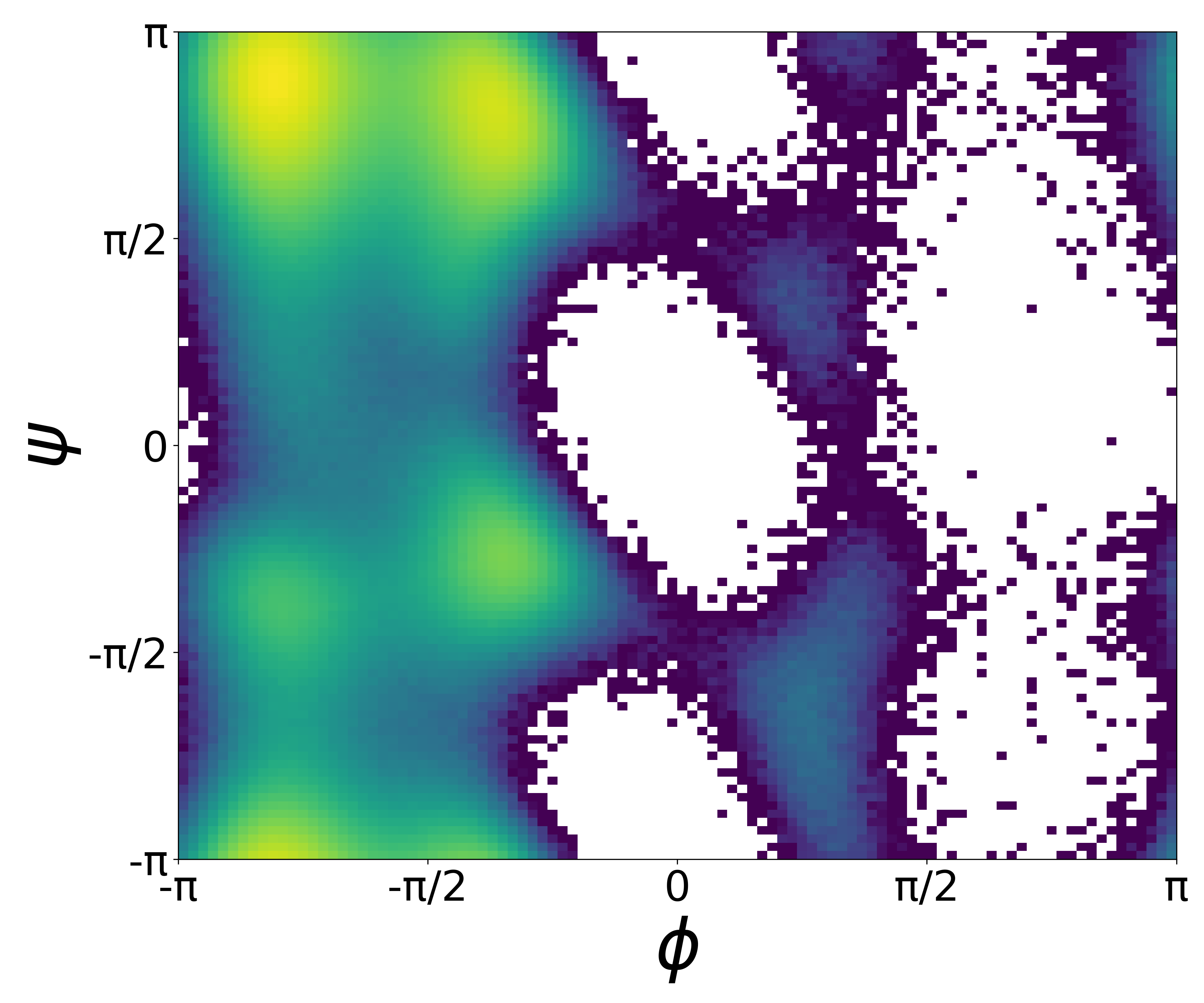}
        \label{fig:aldp-pt}
    }
    \subfigure[\textit{DiGS} ($10^6$ samples)]{\includegraphics[width=0.45 \columnwidth]{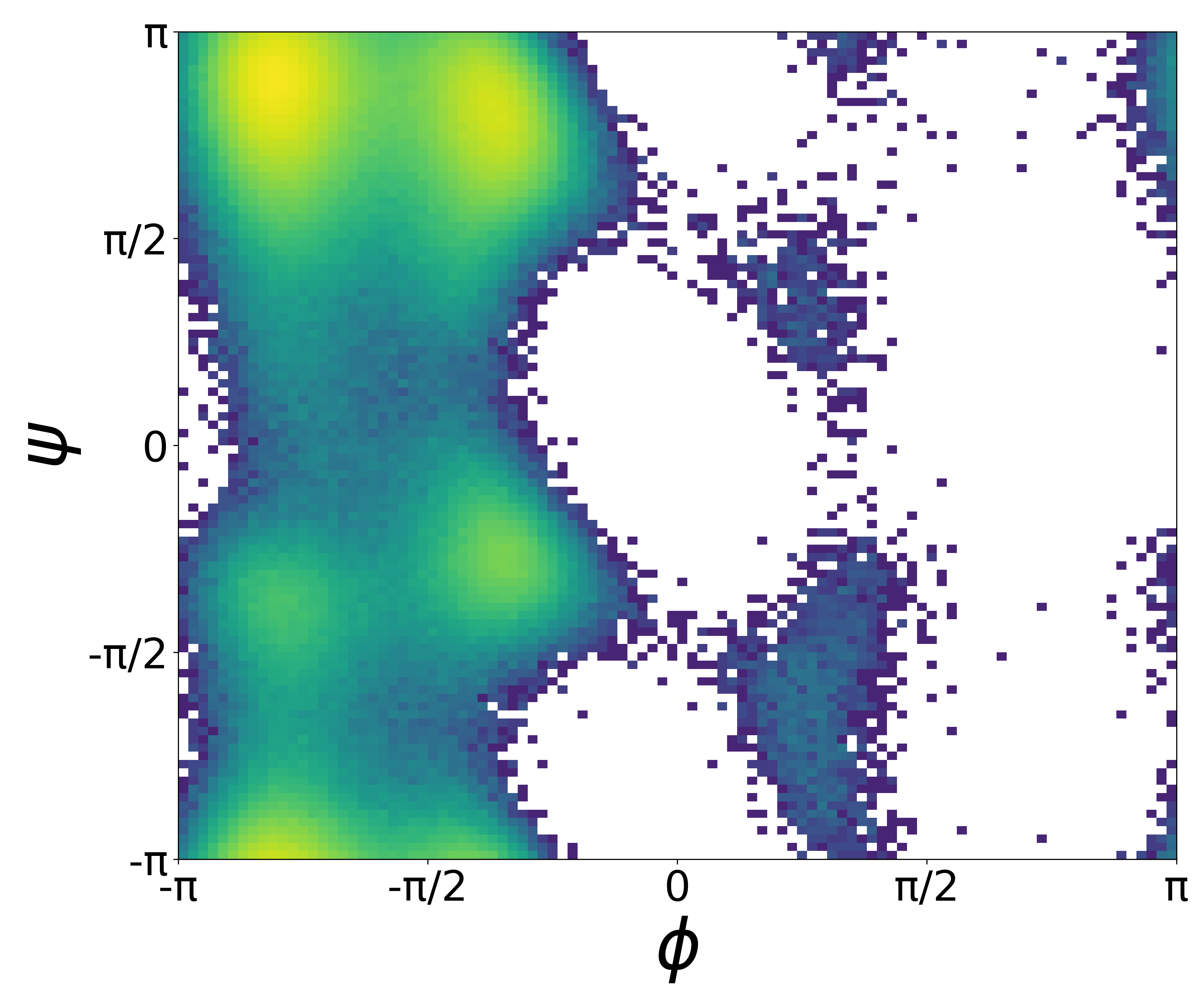}
        \label{fig:aldp-gibbs}
    }
    \caption{Ramachandran plots for alanine dipeptide.}
    \label{fig:ramachandran}
\end{figure}

Each method is initialized at the minimum energy configuration as in~\citet{midgley2023flow} and generates $10^6$ samples of molecular configurations for evaluation. MALA runs 1,000 Langevin steps per sample. HMC runs 1,000 leapfrog steps per sample. DiGS uses the VP schedule with $T=2$ noise levels ($\alpha_2=0.1$ and $\alpha_1=0.9$), each with 100 Gibbs sweeps and 5 MALA denoising sampling steps per Gibbs sweep. Note that PT is the gold-standard method in molecular dynamics (MD) simulation, which serves as the ground-truth. The PT samples are taken from~\citet{midgley2023flow}, which is generated using 21 chains starting at temperature 300K and increasing the temperature by 50K for each subsequent chain, where each chain is constructed by an HMC sampler with 1,000 leapfrog steps per sample. We follow~\citet{midgley2023flow} and treat $10^7$ PT samples as the ground-truth. In addition, we consider a baseline of $10^6$ PT samples as a reference.
The quality of the sampled configurations is assessed by the Ramachandran plot~\citep{ramachandran1963stereochemistry}, which can be used to analyze how the protein folds locally. A Ramachandran plot is a 2D histogram for joint distribution of the two dihedral angles $\phi$ and $\psi$ in the bonds connecting an amino acid to the protein backbone, as shown in Figure~\ref{fig:aldp-visualization}.
Table~\ref{tab:molecule} shows the KL divergences for the Ramachandran plots $p(\phi,\psi)$ and the marginals $p(\phi)$ and $p(\psi)$ between the ground-truth samples and samples generated by each sampler. We can see that DiGS significantly outperforms MALA and HMC with the same number of energy evaluations. Moreover, DiGS also outperforms the gold-standard MD simulation method PT with $23\times$ less energy evaluations. Figure~\ref{fig:ramachandran} shows a visual comparison between the ground-truth Ramachandran plots and the one produced by DiGS, confirming that DiGS captures all modes with the correct weightings.

\section{Conclusion}

Diffusive Gibbs Sampling (DiGS) is an innovative sampler that combines recent development in diffusion models with a novel Metropolis-within-Gibbs scheme. DiGS offers significant improvements in sampling multi-modal distributions, surpassing traditional methods in both efficiency and accuracy. Its applicability across various fields (e.g., Bayesian inference and molecular dynamics) showcases the potential of DiGS to facilitate sampling complex distributions in numerous scientific applications.

\subsection{Limitations and Future Work}
In our experiments, we observed that the acceptance rate of the Metropolis-within-Gibbs scheme given by Equation~\ref{eq:MH:acc:rate} was around $0.15$, due to the random walk behavior of the linear Gaussian proposal as in Equation~\ref{eq:MH proposal}. A potential future work direction would be to design more efficient proposals for the Metropolis-within-Gibbs scheme with higher acceptance rate, for instance, combining the Gaussian convolution kernel and the linear Gaussian proposal by marginalizing out the noisy variable: $q(x|x^{(i-1)})=\int q(x|\tx)p(\tx|x^{(i-1)}) d\tx$ \citep{titsias2018auxiliary}. Empirically, it would be interesting to investigate the the performance of DiGS with different score-based samplers such as HMC in the denoising sampling step in future work. Besides, other multi-modal samplers such as population-based MCMC could be considered as baselines for comparison in future work. We also leave the convergence analysis of DiGS in the general multi-level noise scheduling setting as a future work; see Appendix~\ref{appendix:convergence} for some preliminary discussions.

\clearpage

\section*{Acknowledgements}
We thank Andi Zhang for useful discussions on RDMC, Ruqi Zhang for highlighting Entropy-MCMC, an anonymous reviewer for discussions on PT, and Arnaud Doucet, Michael Hutchinson and Sam Power for pointing out proximal samplers.

MZ and DB acknowledge funding from the Cisco Centre of Excellence. 
WC acknowledges funding via a Cambridge Trust Scholarship (supported by the Cambridge Trust) and a Cambridge University Engineering Department Studentship (under grant G105682 NMZR/089 supported by Huawei R\&D UK). 
JMHL acknowledges support from a Turing AI Fellowship under grant EP/V023756/1.

\section*{Impact Statement}
The goal of this paper is to advance the field of machine learning. There could be many potential positive societal consequences of this methodological work, too numerous to specifically highlight here.

\bibliography{icml2024}
\bibliographystyle{icml2024}

\newpage
\appendix
\onecolumn

\section{Proof of Theorem~\ref{theorem}\label{app:proof}}
\begin{proof}
    A sufficient condition for irreducibility and recurrency of a Markov chain is that the joint distribution $p(x,\tx)$ should satisfy the \emph{positivity condition}~\citep{robert1999monte, roberts1994simple}, which requires $p(x',\tx')>0$ for all $x',\tx'$ such that $p(x')>0$ and $\tilde{p}(\tx')>0$, where $\tilde{p}(\tx')=\int p(\tx'|x)p(x)\dif{x}$. This requirement is satisfied in the DiGS. This is because $p(x,\tx) = p(\tx|x)p(x)$, where the Gaussian convolution kernel $p(\tx|x)$ has full support in $\mathbb{R}^d$. Consequently, if $p(x') > 0$, then it follows that $p(\tx')>0$ and $p(x',\tx') > 0$.
    \vspace{-0.2cm}
\end{proof}
\section{An Analytical Example of Tempering v.s. Convolution}\label{sec:comparison-tempering}

Although tempering-based sampling can alleviate the issue of multi-modal sampling in certain contexts, they still struggle in situations where the modes are less connected or completely isolated.
Consider a toy example of a mixture of two Gaussians in 1D, given by $p(x)=\frac{1}{2}\mathcal{N}(x|\mu,\sigma_g^2)+\frac{1}{2}\mathcal{N}(x|-\mu,\sigma_g^2)$ where both Gaussian components have the same variance $\sigma_g^2$ and symmetric means positioned at $\mu,-\mu$. A point of interest is $x=0$, which lies in the low-density region between the two modes and acts as a barrier point, hindering the state transition from one mode to another.  In this example, the tempered log-density has a closed-form expression up to some constant $C_{\beta}$:
\begin{align}
    \log p_\beta(x=0)
    &=\beta\log p(x=0)\nonumber + C_{\beta}\\
    &=\beta \left(-\frac{\mu^2}{2\sigma_g^2}-\log\sigma_g\right) + C_{\beta}.
\end{align}
For any given inverse temperature $\beta$ and position $\mu$, as $\sigma_g\to0$, we have $\log p_\beta(x=0)\rightarrow-\infty$ and thus $p_\beta(x=0)\rightarrow 0$. This ``Mixture of Deltas'' example shows that tempering methods do not effectively overcome the low-density barrier in such situations. 
Furthermore, if we consider scenarios where each component of the mixture distribution is entirely disconnected, the tempered density in these regions remains zero, as tempering does not alter the support of a distribution.

In contrast, we define the convolved distribution 
$p(\tilde{x})=\int p(\tilde{x}|x)p(x)\dif{x}$ for the same target $p(x)$ with a convolution kernel $p(\tilde{x}|x)=\mathcal{N}(\tilde{x}|x,\sigma^2)$. For any given $\mu$ and $\sigma_g$, without loss of generality, we choose $\sigma\geq \sigma_g/\delta$ with a small constant $\delta>0$. This leads to a lower bound:
\begin{align}
    \log p(\tilde{x}=0)&=
-\frac{\mu^2}{2(\sigma_g^2+\sigma^2)}-\frac{1}{2}\log(2\pi(\sigma_g^2+\sigma^2))\nonumber\\
&\geq -\frac{\mu^2}{2\sigma^2}-\frac{1}{2}\log(2\pi\sigma^2(1+\delta^2)),
\end{align}
illustrating that the convolved log-density at $\tilde{x} = 0$ is lower-bounded for any $\sigma>0$ as $\sigma_g\to0$, since the lower bound of $\log p(\tilde{x} = 0)$ is independent of $\sigma_g$ and remains finite. This approach effectively guarantees the maintenance of a non-negligible density within the bridges connecting different modes of distribution. 
Figure~\ref{fig:mod-intro} presents a one-dimensional comparison between the tempering and convolution methods. 
However, in cases where the standard deviation of the Mixture of Gaussians (MoG) is exceptionally small, the efficacy of tempering diminishes significantly. In such instances, even with a large temperature $T$, tempering fails to connect the modes. Conversely, the convolution method continues to effectively bridge the modes, even with a relatively small $\sigma$. This comparison underscores the convolution method's robustness in handling scenarios with disconnected modes, a situation that also mirrors challenges often encountered in high-dimensional cases, whereas the tempering method struggles. 

\section{Discussions of Convergence Guarantees}\label{appendix:convergence}
In some specific cases, the convergence rate of DiGS may be established. Specifically, the convergence rate of DiGS depends on both the convergence of the inner denoising sampling step and the convergence of the outer Gibbs sampler. For instance, when using MALA to sample from the denoising posterior $p(x|\tx)$, the convergence of MALA has been studied under different conditions such as strong log-concavity and log-Sobolev inequality \citep{vempala2019rapid}. Compared to direct sampling from the target distribution, the L2 regularizer in the log denoising posterior can improve the log-Sobolev condition: 
\begin{align}
    \log p(x|\tx)= -E(x) - \frac{\lVert \alpha x -\tx\rVert^2}{2\sigma^2} + const,
\end{align}
which in turn accelerates the convergence of MALA \citep{RDMC}. Intuitively, this makes the target distribution more ``Gaussian-like''. The convergence of the outer Gibbs sampler can be analyzed following the approach in \citet{chen2022improved}, demonstrating convergence under conditions of strong log-concavity/log-concavity under Wasserstein distance and log-Sobolev inequality under KL divergence. Therefore, the theoretical convergence guarantee of DiGS with a single noise level can be obtained by combining the convergence analyses of the inner MALA denoising sampling step and the outer Gibbs sampler from \citet{vempala2019rapid,chen2022improved,RDMC}.

However, in addition to the aforementioned factors, the multi-level noise schedule also plays a part in accelerating the convergence speed. Furthermore, our empirical investigation revealed that the performance of DiGS for practical problems such as molecular dynamics could also be affected by the initial condition (e.g., we followed the convention of molecular dynamics to initialize DiGS at the minimum energy configuration, which turns out to be better than random initialization). These analyses are out of the scope of this work, and we leave the theoretical study of the general DiGS framework as a future research direction.

\section{Experimental Details~\label{app:exp}}
In all experiments, the step sizes of MALA and HMC are tuned via trial-and-error so that the acceptance rates are close to optimal values $0.574$~\citep{roberts1998optimal} and $0.65$~\citep{neal2011mcmc}, respectively.

\subsection{Comparison of Initialization Strategies for the Denoising Sampling Step\label{app:init-comparison}}
For the comparison of three initialization strategies for the denoising sampling step in Figure~\ref{fig:unequal:mog} and Table~\ref{tab:mmd:unequal:mog} in Section \ref{sec:init-denoising-sampling}, we run DiGS with $\alpha=1, \sigma=1$ for 200 Gibbs sweeps on an MoG with unbalanced weights. The denoising sampling step begins with an initialization using the Metropolis-Hastings (MH) algorithm, followed by 50 MALA steps with a step size of $1{\times}10^{-3}$. For evaluation, we employ the Maximum Mean Discrepancy (MMD) \citep{gretton2012kernel} that utilizes 5 kernels with bandwidths $\{2^{-2}, 2^{-1}, 2^{0}, 2^{1}, 2^{2}\}$.

\subsection{Comparison of Gaussian Convolution Hyperparameters\label{app:conv-param-comparison}}
For the convolution parameter comparison in Section~\ref{sec:choose:parameter}, we use DiGS to generate 1,000 samples. For each sample, we execute 1,000 Gibbs sweeps. The denoising sampling step begins with an initialization using the Metropolis-Hastings (MH) algorithm, followed by 10 MALA steps. The step size for MALA is set at $1 \times 10^{-3}$. We set $\sigma=1.0$ and vary $\alpha$ across $\{0.01, 0.05, 0.1, 0.2, 0.3, 0.4, 0.5, 0.6, 0.7, 0.8, 0.9, 1.0, 2.0, 5.0\}$ for the experiment shown in Figure~\ref{fig:effects:alpha}. Similarly, we set $\alpha=1.0$ and vary $\sigma$ across $\{0.1, 0.3, 0.5, 1, 2, 3, 4, 5, 6, 7, 8, 10, 15, 20\}$ for the experiment shown in Figure~\ref{fig:effects:std}. For evaluation, we employ the Maximum Mean Discrepancy (MMD) \citep{gretton2012kernel} that utilizes 5 kernels with bandwidths $\{2^{-2}, 2^{-1}, 2^{0}, 2^{1}, 2^{2}\}$.

\subsection{Multi-Level Noise Scheduling\label{app:multi-level}}
For the multi-level noise experiment with the VP schedule in Section~\ref{sec:multi-level}, we define $\alpha_t = \alpha_{T} + (T - t) \Delta_\alpha$, where $\Delta_\alpha = (\alpha_1 - \alpha_{T}) / (T - 1)$ and $\sigma_t = \sqrt{1 - \alpha_t^2}$. We set $\alpha_1 = 0.9$ and $\alpha_T = 0.1$, and experiment with $T \in \{2, 3, 4, 5\}$ for the comparison shown in Figure~\ref{fig:effects:multilevel}. For evaluation, we employ the Maximum Mean Discrepancy (MMD) \citep{gretton2012kernel} that utilizes 5 kernels with bandwidths $\{2^{-2}, 2^{-1}, 2^{0}, 2^{1}, 2^{2}\}$.

\subsection{Comparison with Parallel Tempering\label{app:comparison-pt}}
For comparison with parallel tempering on the ``Mixture of Delta'' problem in Figure~\ref{fig:mod} in Section~\ref{sec:tempering-intro}, each sampler is initialized at the origin and generates 1,000 samples. DiGS employs a single noise level Gaussian convolution with parameters $\alpha=1$ and $\sigma=1$, complemented by 1,000 Gibbs sweeps per sample. The denoising sampling step in each Gibbs sweep begins with MH initialization, followed by 5 steps of MALA with a step size of $1{\times}10^{-3}$. PT consists of 5 chains with temperatures $\tau{=}\{1.0, 5.62, 31.62, 177.83, 1000.0\}$, where each chain is constructed by an HMC sampler with 1,000 leapfrog steps per sample and a step size of $1{\times}10^{-2}$.

To make PT work on this problem, one needs to use 20 chains with temperatures $\tau{=}\{$1.0, 2.07, 4.28, 8.86, 18.33, 37.93, 78.48, 162.38, 335.98, 695.19, 1432.45, 2976.35, 6158.48, 12742.75, 26366.51, 54555.95, 112883.79, 233572,15, 483293.02, 1000000.0$\}$ and the DEO scheme \citep{deng2023non}, where each chain is constructed by an HMC sampler with 500 leapfrog steps per sample and a step size of $1{\times}10^{-2}$.

\subsection{Comparison with RDMC\label{app:comparison-rdmc}}
For the comparison with Reverse Diffusion Monte Carlo (RDMC) in Figure~\ref{fig:digis_rdmc} in Section~\ref{sec:rdmc}, we follow the original setting discussed in~\citet{RDMC} and use ULA for generating samples for the posterior distribution. The score function of the reverse distribution using $K=5$ to construct the score approximation
\begin{equation}
\nabla_{x_t}\log p(x_t) \approx \left(\frac{\alpha}{K}\sum_{k=1}^K x_0^{(t,k)}-x_t\right)/\sigma_t^2,
\end{equation}
where $x_0^{(t,k)} \sim p(x_0|x_t)$ denotes the samples drawn using ULA with $L_{ULA}=5$ steps and a step size of $1 {\times}10^{-2}$. These samples are initialized through importance sampling (IS) with $S_{is}=100$ samples. The discretization step size $\eta$ is set to $\Gamma/T$, with the scaling factor $a_t = e^{-(\Gamma-t \cdot \eta)}$ and the standard deviation $\sigma_t = \sqrt{1-\alpha_t^2}$. Following the procedure in~\citet{RDMC}, we further apply ULA for additional $L_{ULA}$ steps after RDMC. The parameter $\Gamma$ is fixed at $0.1$ for our experiments, and we explore different values of $T \in \{1,2,3,4\}$, necessitating $TK(L_{ULA}+S_{is})+L_{ULA}$ energy evaluations to generate a single sample. 

For DiGS, we use DiGS with one noise level Gaussian convolution with parameter $\alpha=1,\sigma=1$. For sampling from the denoising distribution, we use the MH initialization followed by the ULA sampling with $L_{ULA}=5$ steps and a step size of $1{\times}10^{-2}$. This MH+ULA scheme ensures a fair comparison to the IS+ULA scheme used in the RDMC. 
We vary $S_{gibbs}\in \{1,2,3,4,5,6,7,8,9,10\}$ Gibbs sweeps to generate a sample, which takes $S_{gibbs}(L_{ULA}+2)$ energy evaluations in total, where the constant $2$ accounts for the two energy evaluations required by MH. For both DiGS and RDMC, a total of 1,000 samples are generated for comparison.

For evaluation, we employ the Maximum Mean Discrepancy (MMD) \citep{gretton2012kernel} that utilizes 5 kernels with bandwidths $\{2^{-2}, 2^{-1}, 2^{0}, 2^{1}, 2^{2}\}$.

\subsection{Mixture of 40 Gaussians}
For the MoG-40 experiment in Section~\ref{sec:mog-40}, each method is initialized at the origin and generates $10^4$ samples for evaluation. MALA runs 1,000 Langevin steps per sample with a step size of $1 \times 10^{-1}$. HMC runs 1,000 leapfrog steps per sample with a step size of $1 \times 10^{-1}$. PT consists of 5 chains with temperatures $\tau=\{1.0, 5.62, 31.62, 177.83, 1000.0\}$, where each chain is constructed by an HMC sampler with 200 leapfrog steps per sample with a step size of $1 \times 10^{-1}$. DiGS uses $T=1$ noise level with $\alpha=0.1$ and $\sigma^2=1-\alpha^2$, 200 Gibbs sweeps, and 5 MALA denoising sampling steps per Gibbs sweep with a step size of $1 \times 10^{-1}$.

\subsection{Bayesian Neural Networks}
For the BNN experiment in Section~\ref{sec:bnn}, all methods are initialized at the same random sample from the prior and generate $150$ samples from the posterior $p(\theta|D_{\text{train}})$ for evaluation. MAP runs $7.5\times 10^5$ full-batch gradient descent steps with a step size of $3 \times 10^{-2}$. MALA runs 5,000 Langevin steps per sample with a step size of $1 \times 10^{-4}$. HMC runs 5,000 leapfrog steps per sample with a step size of $5 \times 10^{-4}$. PT consists of 5 chains with temperatures $\tau=\{1.0, 5.62, 31.62, 177.83, 1000.0\}$, where each chain is constructed by an HMC sampler with 1,000 leapfrog steps per sample with a step size of $5 \times 10^{-4}$. DiGS uses the VP schedule with $T=5$ noise levels, ranging from $\alpha_T=0.1$ to $\alpha_1=0.9$, each with 100 Gibbs sweeps and 10 MALA denoising sampling steps per Gibbs sweep with a step size of $1 \times 10^{-4}$.

\subsection{Molecular Configuration Sampling}\label{appendix:addition-mol}
For the molecular configuration sampling experiment in Section~\ref{sec:molecular}, each method is initialized at the minimum energy configuration as in~\citet{midgley2023flow} and generates $10^6$ samples of configurations for evaluation. MALA runs 1,000 Langevin steps per sample with a step size of $1 \times 10^{-4}$. HMC runs 1,000 leapfrog steps per sample with a step size of $1 \times 10^{-3}$. DiGS uses the VP schedule with $T=2$ noise levels ($\alpha_2=0.1$ and $\alpha_1=0.9$), each with 100 Gibbs sweeps and 5 MALA denoising sampling steps per Gibbs sweep with a step size of $1 \times 10^{-4}$. Note that PT is the gold-standard method in molecular dynamics (MD) simulation, which serves as the ground-truth. The PT samples are taken from~\citet{midgley2023flow}, which is generated using 21 chains starting at temperature 300K and increasing the temperature by 50K for each subsequent chain, where each chain is constructed by an HMC sampler with 1,000 leapfrog steps per sample. We follow~\citet{midgley2023flow} and treat $10^7$ PT samples as the ground-truth. In addition, we consider a baseline of $10^6$ PT samples as a reference.

\begin{table}[t]
    \normalsize
    \caption{Sample quality on a 200-dimensional mixture of Gaussians problem.}
    \label{tab:mog-200d}
        \centering
        \begin{tabular}{ccc}
            \toprule
            Sampler & MMD & \#energy evaluations \\
            \midrule
            MALA & $0.123\pm0.035$ & $10^6$ \\
            HMC & $0.104\pm0.025$ & $10^6$ \\
            PT & $0.094\pm0.014$ & $10^6$ \\
            \textit{DiGS} & $\mathbf{0.027\pm0.010}$ & $10^6$  \\
            \bottomrule
        \end{tabular}
\end{table}

\section{Additional Experimental Results}

\subsection{200-Dimensional Mixture of Gaussians}
We take the MoG target in Figure~\ref{fig:unbalanced:mog:density} and generalize it to the 200-dimensional space. For each compared sampler, we draw 1,000 samples within a given compute budget of $10^6$ energy evaluations. All samplers are tuned to achieve their optimal acceptance rate as described in Section~\ref{sec:experiments}. Table~\ref{tab:mog-200d} shows that DiGS significantly outperforms all the other samplers even in the high-dimensional space.

\subsection{Molecular Configuration Sampling}
\begin{figure}[t]
    \centering
    \subfigure[Dihedral angles $\phi,\psi$ in alanine dipeptide]{
    \raisebox{8ex}{\includegraphics[width=0.43\linewidth]{./img/aldp/3d-mol.png}}
    \label{fig:aldp-visualization-appendix}
}
    \subfigure[MD ($10^7$ samples, $2.3{\times}10^{11}$ energy evaluations)]{\includegraphics[width=0.43\linewidth]{./img/aldp/md7.png}
        \label{fig:aldp-gt-appendix}
    }
    \subfigure[MALA ($10^6$ samples, $1.0{\times}10^{9}$ energy evaluations)]{\includegraphics[width=0.43\linewidth]{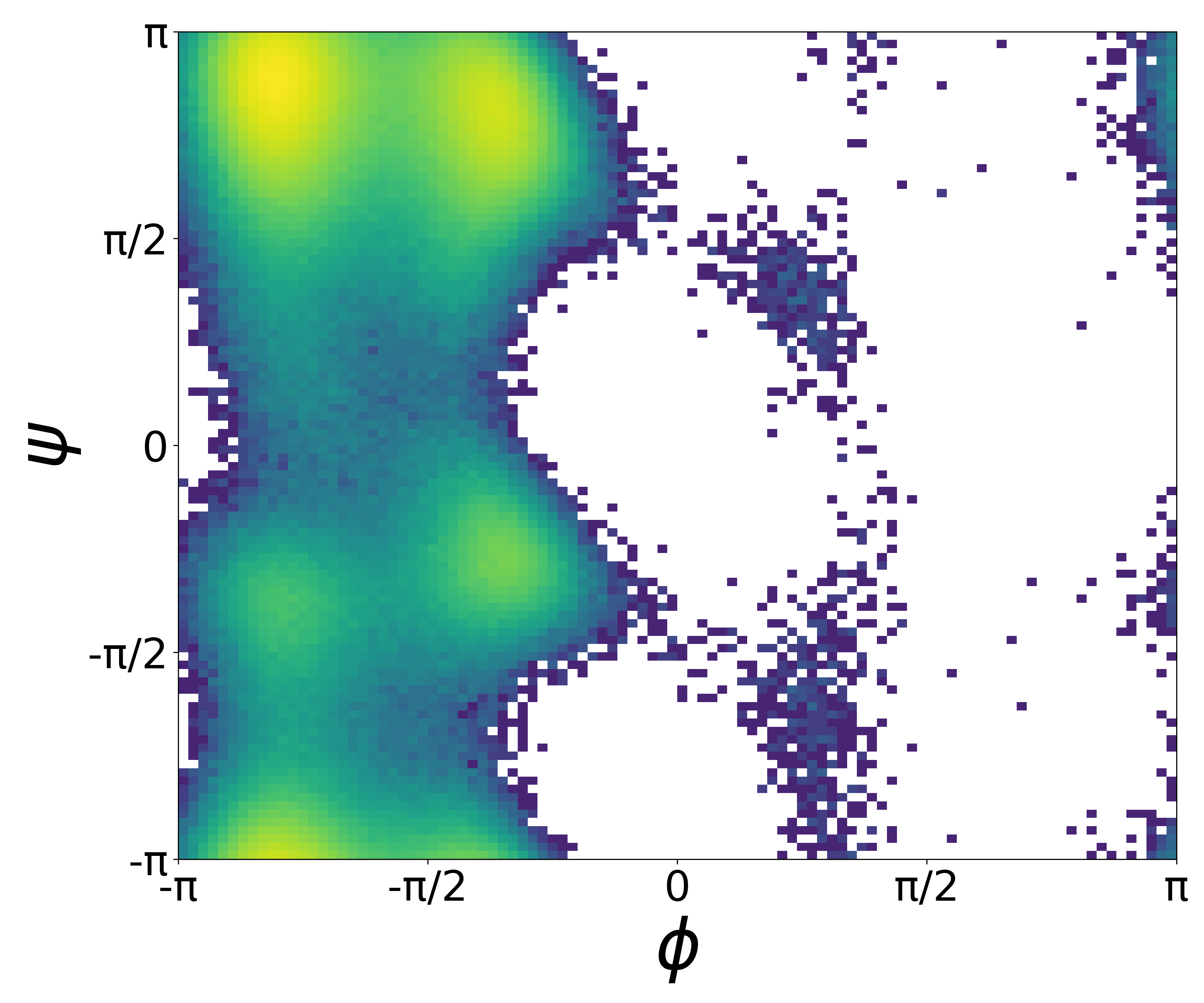}
        \label{fig:aldp-lg-appendix}
    }
    \subfigure[HMC ($10^6$ samples, $1.0{\times}10^{9}$ energy evaluations)]{\includegraphics[width=0.43\linewidth]{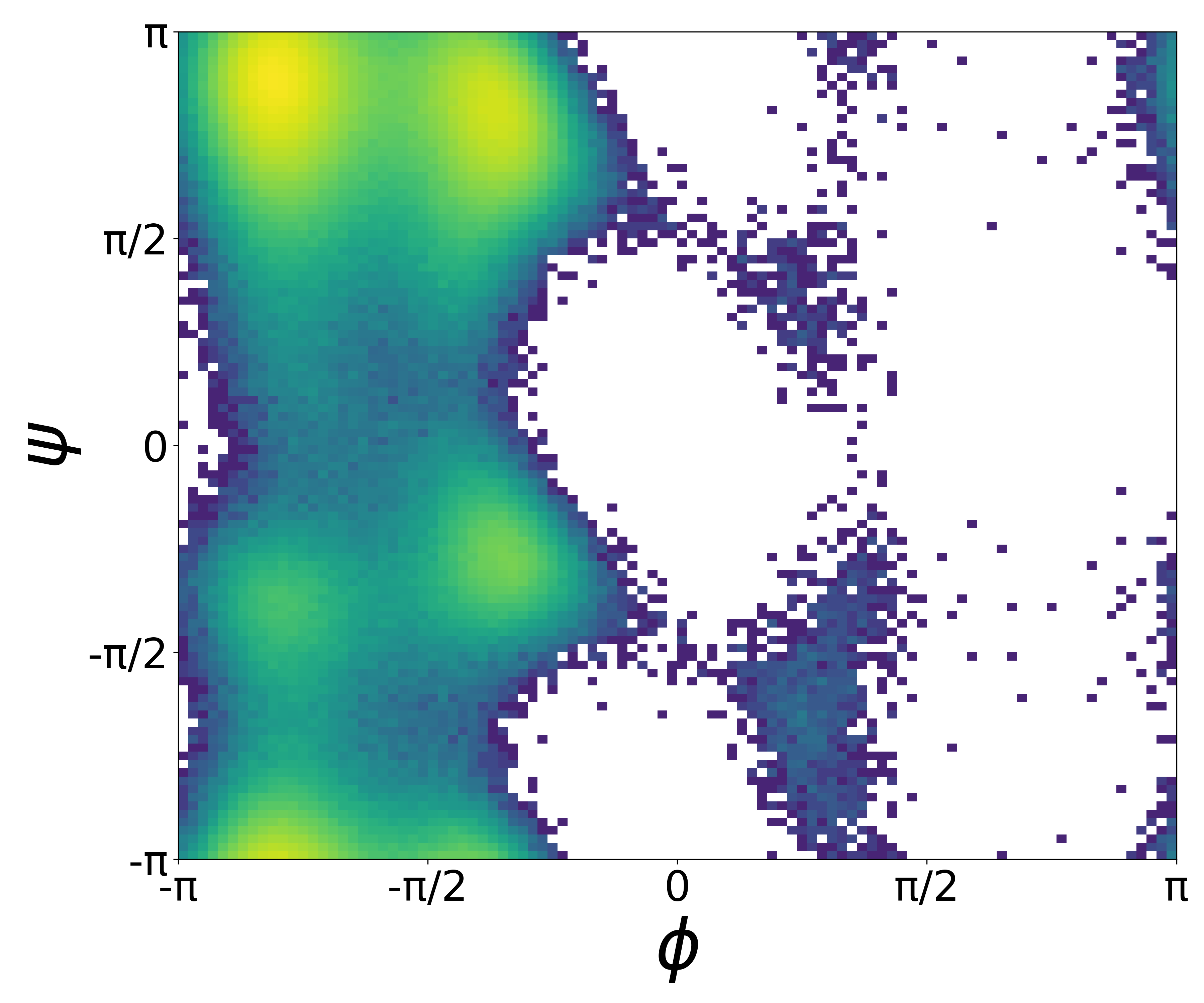}
        \label{fig:aldp-hmc-appendix}
    }
    \subfigure[PT ($10^6$ samples, $2.3{\times}10^{10}$ energy evaluations)]{\includegraphics[width=0.43\linewidth]{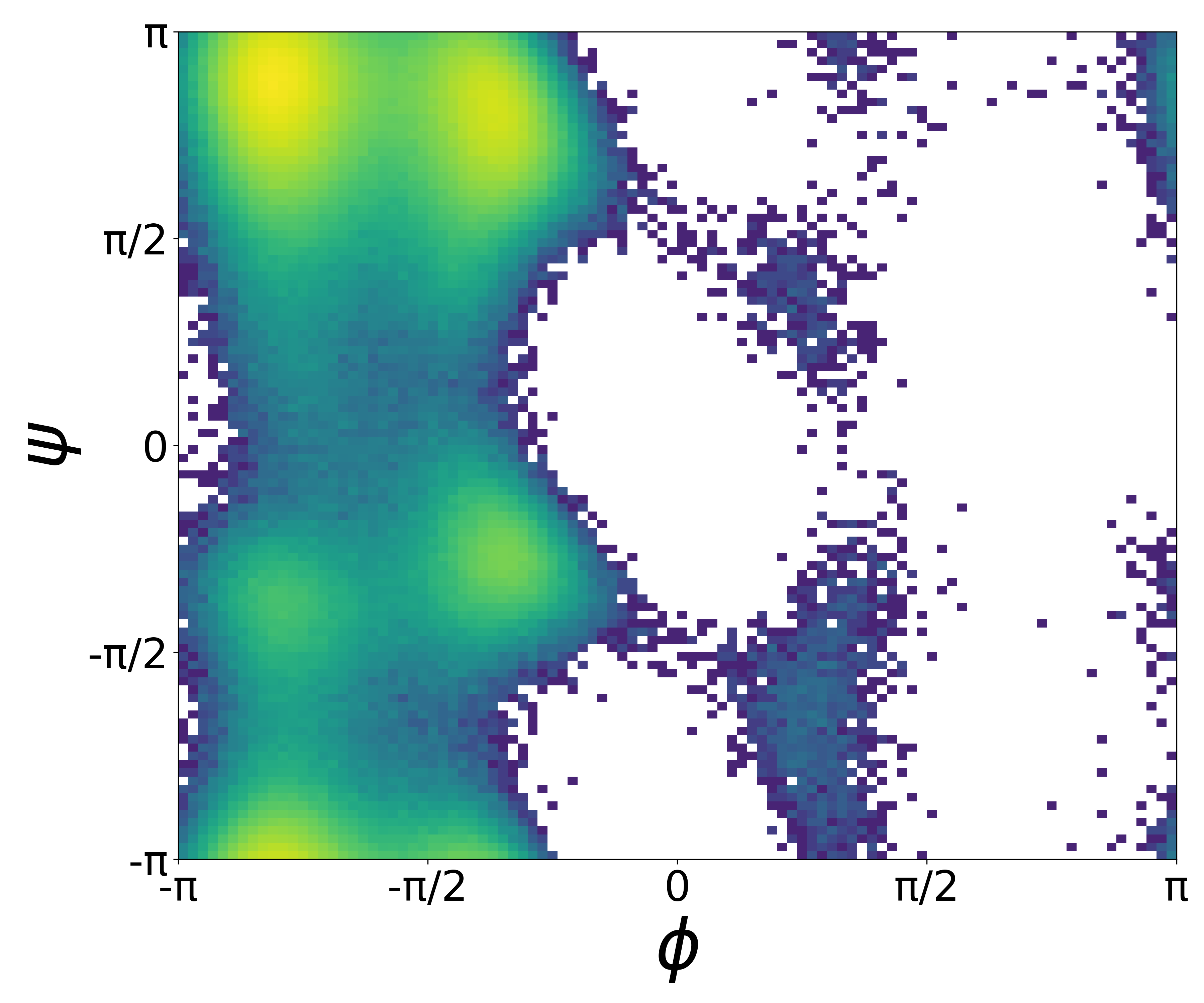}
        \label{fig:aldp-pt-appendix}
    }
    \subfigure[\textit{DiGS} ($10^6$ samples, $1.0{\times}10^{9}$ energy evaluations)]{\includegraphics[width=0.43\linewidth]{./img/aldp/gibbs.png}
        \label{fig:aldp-gibbs-appendix}
    }
    \caption{(a) Visualization of the dihedral angles $\phi$ and $\psi$ in alanine dipeptide \citep{midgley2023flow}. (b)-(d) Ramachandran plots for the dihedral angles $\phi$ and $\psi$ of alanine dipeptide. MD simulation (PT with $10^7$ samples) serves as ground-truth.}
    \label{fig:ramachandran-appendix}
\end{figure}

The Ramachandran plots produced by all compared methods for the molecular configuration sampling experiment in Section~\ref{sec:molecular} can be found in Figure~\ref{fig:ramachandran-appendix}.

\end{document}